\documentclass[11pt]{article}
\usepackage[authoryear,longnamesfirst]{natbib}
\usepackage[a4paper,margin=1in]{geometry}

\usepackage{algorithm}
\usepackage{algorithmic}
\usepackage{array}
\usepackage{textcomp}
\usepackage{url}
\usepackage{verbatim}
\usepackage{graphicx}
\usepackage{amsmath}
\usepackage{tabularx}
\usepackage{multirow}
\usepackage{booktabs}
\usepackage{subcaption}
\usepackage{makecell}
\usepackage{float}
\usepackage{amsfonts}
\usepackage{amssymb}
\usepackage{amsthm}
\usepackage{comment}
\usepackage[table]{xcolor}
\usepackage{bm}

\title{A Learnable Wavelet Transformer for Long-Short Equity Trading and Risk-Adjusted Return Optimization}
\author{
Shuozhe Li\thanks{Equal contribution.}\\
The University of Texas at Austin\\
\texttt{shuozhe.li@utexas.edu}
\and
Du Cheng\footnotemark[1]\\
Northeastern University\\
\texttt{v6hit7cd@gmail.com}
\and
Amy Zhang\\
The University of Texas at Austin\\
\texttt{amy.zhang@austin.utexas.edu}
\and
Leqi Liu\\%\thanks{Corresponding author.}\\
The University of Texas at Austin\\
\texttt{leqiliu@utexas.edu}
}
\date{}

\begin{document}

\maketitle

\begin{abstract}
Learning profitable intraday trading policies from financial time series is challenging due to heavy noise, non-stationarity, and strong cross-sectional dependence among related assets. We propose \emph{WaveLSFormer}, a learnable wavelet-based long-short Transformer that jointly performs multi-scale decomposition and return-oriented decision learning. Unlike standard time-series forecasting that optimizes prediction error and typically requires a separate position-sizing or portfolio-construction step, our model directly outputs a market-neutral long/short portfolio and is trained end-to-end on a trading objective with risk-aware regularization. Specifically, a learnable wavelet front-end generates low-/high-frequency components via an end-to-end trained filter bank, guided by spectral regularizers that encourage stable and well-separated frequency bands. To fuse multi-scale information, we introduce a low-guided high-frequency injection (LGHI) module that refines low-frequency representations with high-frequency cues while controlling training stability. The model outputs a portfolio of long/short positions that is rescaled to satisfy a fixed risk budget and is optimized directly with a trading objective and risk-aware regularization. Extensive experiments on five years of hourly data across six industry groups, evaluated over ten random seeds, demonstrate that WaveLSFormer consistently outperforms MLP, LSTM and Transformer backbones, with and without fixed discrete wavelet front-ends. On average in all industries, WaveLSFormer achieves a cumulative overall strategy return of $0.607 \pm 0.045$ and a Sharpe ratio of $2.157 \pm 0.166$, substantially improving both profitability and risk-adjusted returns over the strongest baselines.
\end{abstract}


\section{Introduction}
\label{sec:introduction}
% Start the introduction with a drop cap using the IEEEPARstart command.
Financial economists have long debated whether asset prices fully reflect all available information. 
The Efficient Market Hypothesis (EMH) suggests that, under rational expectations and frictionless markets, risk-adjusted excess returns should be unpredictable~\cite{fama1970efficient}. 
Nevertheless, decades of empirical evidence indicate that certain technical rules and cross-sectional patterns can yield economically meaningful profits in practice~\cite{brock1992simple}. 
These observations motivate systematic methods for extracting tradable signals from historical prices and related market features.

With modern computing and large historical datasets, deep learning has become a central tool for modeling complex dependencies in financial time series. 
A wide range of architectures, from deep feed-forward networks to recurrent models, have been explored for stock prediction and trading~\cite{kohzadi1996comparison,ozbayoglu2020deep,zhang2024dlreview,giantsidi2025dlreview}. 
Despite promising results, a key challenge remains: many pipelines are formulated as \emph{forecasting} problems optimized by point-wise prediction losses, whereas real trading performance is determined by sequential \emph{position decisions} and their accumulated P\&L over time. 
This objective mismatch can produce models that appear strong under regression/classification metrics yet fail to deliver robust out-of-sample strategy returns, especially when signals are weak and regimes shift.

Financial time series further complicate end-to-end learning. 
Returns are noisy, heavy-tailed, and heteroskedastic, and exploitable structure may only be visible at certain temporal scales. 
Moreover, assets within the same market segment are not independent: co-movements and lead-lag effects are prevalent, so effective decisions often require jointly modeling a group of related assets rather than treating each series in isolation~\cite{ozbayoglu2020deep,giantsidi2025dlreview}. 
These characteristics motivate models that can extract stable multi-scale representations while aligning training with trading-oriented objectives.

Wavelet transforms are a natural candidate for multi-scale analysis because they provide a time-frequency decomposition that separates slow-moving trends from transient fluctuations. 
Prior studies often apply fixed discrete wavelet transforms as preprocessing before downstream predictors~\cite{peng2021multiresolution,rao2022wavelet,zhang2024wtarimalstm,ziolkowski2025dwtlstm}, and wavelet-integrated deep architectures have been surveyed more broadly~\cite{wu2025waveletintegrated}. 
In parallel, Transformers have gained popularity for time-series modeling, and wavelet-Transformer hybrids have been proposed to enhance forecasting by combining multi-scale decomposition with attention~\cite{elabid2022wtransformers,wei2025wavelettransformer,ping2025waveletenhanced}. 
However, for trading, two limitations are particularly relevant: many hybrids still rely on \emph{fixed} wavelets and emphasize forecasting accuracy rather than trading performance, and cross-frequency interactions are often handled implicitly without explicitly controlling how high-frequency information should refine low-frequency representations for stable decision learning.

To address these gaps, we propose \emph{WaveLSFormer}, a trading policy model that outputs long/short positions for a group of related assets and is trained end-to-end with a trading-aware objective. 
WaveLSFormer couples a \emph{learnable} wavelet front-end with a Transformer backbone: the wavelet module is optimized jointly with the downstream objective to learn task-adaptive frequency bands, and we introduce a gated cross-frequency injection mechanism that injects high-frequency cues into the low-frequency branch aiming to filter the noise. 
We further apply a risk-budget rescaling step so that the predicted positions satisfy a fixed risk budget, making the learning target closer to practical portfolio construction. 
Our main contributions are as follows:
{\renewcommand{\labelenumi}{\arabic{enumi})}
% Override enumeration and use the standard enumerate environment
\begin{enumerate}\itemsep0pt \parskip0pt \parsep0pt
  \item We introduce \emph{WaveLSFormer}, which replaces fixed wavelet preprocessing with an end-to-end learnable wavelet filter bank,
  coupled with a Transformer backbone for intraday long/short trading.
  \item We propose a stable cross-frequency fusion strategy via low-guided high-frequency injection with gated residual control,
  and integrate risk-budget rescaling to produce practical long/short positions.
  \item We conduct extensive experiments on five years of hourly U.S.\ equity data across multiple industries, and show that WaveLSFormer improves the overall ROI from $0.225\pm0.056$ to $0.607\pm0.045$ and Sharpe from $1.024\pm0.122$ to $2.157\pm0.166$, while also improving over the best wavelet-enhanced LSTM from $0.317\pm0.050$ to $0.607\pm0.045$ in ROI and from $1.879\pm0.158$ to $2.157\pm0.166$ in Sharpe.
\end{enumerate}}

\begin{comment}
The remainder of this paper is organized as follows. 
Section~\ref{sec:related-work} reviews related work. 
Section~\ref{sec:problem-formulation} formulates the problem. 
Section~\ref{sec:data-construction-pipeline} introduces the process of constructing data prior to training model. 
Section~\ref{sec:architecture} shows the model design details.
Section~\ref{sec:training-design} describes partition loss functions and the final whole loss function.
Section~\ref{sec:experiment} presents experimental settings and results. 
Section~\ref{sec:conclusion} concludes the paper with a discussion of limitations and future directions.
\end{comment}

\section{Related Work}
\label{sec:related-work}
\subsection{Forecasting Models and Trading Objectives}
Deep learning has been widely adopted for financial time series due to its ability to capture nonlinear dynamics, regime shifts, and complex temporal dependencies. Representative applications include CNN/LSTM-style predictors for stock forecasting~\cite{HOSEINZADE2019273,Nelson2017LSTM,Moghar2020LSTM,Bao2017DLF,buczynski2023dlfts}. 
However, a recurring challenge is the objective mismatch between training and deployment. Many methods optimize point-wise forecast losses like MSE or MAE, while trading performance depends on sequential position decisions, constraints, and risk exposure. As a result, sequence predictors do not necessarily intend to achieve robust strategies~\cite{krauss2017deep}. 
Transformers~\cite{vaswani2017attention} have also become a dominant sequential modeling architecture and have been adapted to time-series forecasting, with influential variants such as Informer~\cite{zhou2021informer} and increasingly use in financial prediction~\cite{Mozaffari2024Transformer,Kabir2025LSTMTrans}. 
Yet many Transformer-based financial studies still evaluate primarily under forecasting metrics and then apply external trading rules to obtain positions, which can weaken end-to-end alignment with economic utility.

\subsection{Wavelet Transformer and Frequency Fusion}
Financial signals are noisy and non-stationary, motivating wavelet transforms as a time-frequency tool for denoising and multi-scale decomposition before prediction. Classic wavelet neural hybrids and wavelet-reconstruction pipelines, including combinations with ARIMA/LSTM or related deep models, have been explored to improve forecast accuracy on volatile markets~\cite{Lei2018WNN,Peng2021WaveletDL,peng2021multiresolution,rao2022wavelet,zhang2024wtarimalstm,ziolkowski2025dwtlstm,kumar2016dwtann}, and wavelet features have also been used in trading-oriented settings such as reinforcement learning~\cite{lee2021waveletdrl}. 
More recently, wavelet Transformer hybrids construct multi-scale inputs and use attention to aggregate patterns across time and scales~\cite{elabid2022wtransformers,wei2025wavelettransformer,ping2025waveletenhanced}. In finance, Stockformer~\cite{ma2025stockformer} is a recent attempt to integrate Transformer-style modeling with trading-oriented outputs, and surveys summarize wavelet-integrated deep networks for noisy time series~\cite{wu2025waveletintegrated}. 
A key modeling question is how information should flow across frequencies: simple concatenation or implicit fusion can allow unstable high-frequency components to dominate learning under noisy financial signals. 
In this work, we address both the objective mismatch and the frequency-fusion challenge by learning an end-to-end wavelet filter bank jointly with a trading-oriented objective and adopting a low-guided high-frequency injection mechanism, where attention is computed from the low-frequency branch and high-frequency values are injected through a gated residual pathway for stable fusion. Empirically, we benchmark carefully matched MLP/LSTM/Transformer backbones with and without fixed or learnable wavelet front-ends under the same trading protocol.

\section{Problem Formulation}
\label{sec:problem-formulation}
In this section we formalize the trading task addressed by our model. We begin by specifying the representation of multi-asset price series and returns, then describe the decision times, model inputs and outputs, and finally state the ideal trading objective that motivates our training loss and evaluation metrics.

\subsection{Input Windows and Model Outputs}
\noindent
We consider the instruments in a given industry, including related ETFs and sector indices. 
We index instruments by $j\in\{1,\dots,d\}$ and hourly bars by $t\in\{1,\dots,T\}$. 
For instrument $j$, let $Open_{j,t}$ and $Close_{j,t}$ denote the open and close prices at time $t$. 
The simple return and log return of bar $t$ are
\begin{equation}
r_{j,t} := \frac{Close_{j,t}}{Open_{j,t}} - 1,
\qquad
\ell_{j,t} := \log(1+r_{j,t}).
\label{eq:price_return}
\end{equation}

In our experiments, we use the log price return $\ell_{j,t}$ defined in Eq.~\eqref{eq:price_return} and collect
\begin{equation*}
\mathbf r_t=(r_{1,t},\ldots,r_{d,t})^\top,\qquad
\boldsymbol{\ell}_t=(\ell_{1,t},\ldots,\ell_{d,t})^\top\in\mathbb{R}^d .
\end{equation*}

We adopt a rolling-window formulation: at decision time $t$ the model observes
\begin{equation*}
  \mathbf{X}_t
  =
  \bigl[\boldsymbol{\ell}_{t-L+1},\dots,\boldsymbol{\ell}_{t}\bigr]
  \in \mathbb{R}^{d\times L},
\end{equation*}
with columns ordered from oldest to most recent. Additional channels are absorbed into $d$.

Given $\mathbf{X}_{t-1}$, the model outputs a scalar trading signal $p_t = f_{\theta}(\mathbf{X}_{t-1})\in\mathbb{R}$, where $f_\theta$ is WaveLSFormer. We map $p_t$ to a bounded position via
\begin{equation}
  w_t = 2\sigma(p_t)-1 = \tanh\!\left(\frac{p_t}{2}\right),
  \label{eq:tanh-position}
\end{equation}
so $w_t>0$ corresponds to long and $w_t<0$ corresponds to short. $|w_t|$ controls position magnitude.

As discussed in Section~\ref{subsec:pos-rule}, using $w_t$ directly can confound signal quality with aggressiveness. 
We therefore apply a global risk-budget scaling calibrated on the validation set and define the executable position as $\tilde{w}_t = h(w_t)\in[\tilde{w}_{\min},\tilde{w}_{\max}]$, where $h(\cdot)$ is a fixed piecewise-linear mapping that enforces a prescribed average leverage and clips extremes. 
In long-short experiments we set $\tilde{w}_{\min}=-L$ and $\tilde{w}_{\max}=L$, while in long-only or short-only settings one bound is set to zero. 
Section~\ref{subsec:pos-rule} gives the exact form of $h(\cdot)$.

\subsection{Return Representation}
\label{sec:return_representation}
In each experiment we select one target asset $j^\star \in \{1,\dots,d\}$ whose trading performance we evaluate, and treat the remaining instruments as auxiliary series providing contextual information. This reflects a common use case in industry-level trading, where a trader is primarily interested in a particular stock or ETF, but leverages signals from other stocks within the same sector. The single-period realized return of the trading strategy between $t$ and $t+1$ is then 
\begin{equation}
R_{t+1} \;=\;  1 + \tilde{w}_t\! \cdot\! r_{j^\star,t+1}
\label{eq:single_return}
\end{equation}

The return on investment (ROI) of the trading strategy over an evaluation period $\mathcal{T}$ is
\begin{equation}
\label{eq:roi}
  {ROI}_\mathcal{T}
  \;=\;
  \exp\bigl( \sum_{t \in \mathcal{T}} \log\!\left( R_{t} \right) \bigr) - 1.
\end{equation}

\subsection{Optimization Target}
From a trading perspective, the goal is not to minimize a point-wise prediction error, but to control the distribution of realized strategy returns. In our setting the single–period return of the strategy between $t$ and $t+1$ is Eq.~\eqref{eq:single_return} and over an evaluation horizon $\mathcal{T}$, the corresponding ROI is Eq.~\eqref{eq:roi}.

Maximizing return alone is not sufficient in practice, since extremely volatile strategies with large draw-downs are usually unacceptable. 
A standard risk-adjusted performance measure in finance is the annualized Sharpe ratio, defined on simple returns $R_t$ as
\begin{equation}
\mathcal{S}_{\mathrm{ann}}
=\sqrt{K}\,
\frac{\mathbb{E}\!\left[R_t-r_f/K\right]}
{\sqrt{\mathrm{Var}\!\left(R_t-r_f/K\right)}},
\qquad
K \approx 252\times 6.5,
\label{eq:sharpe_def}
\end{equation}
where $r_f$ is the annual risk-free rate.
In our hourly–bar backtests, $R_t$ is already a small return per bar. For model comparison on a fixed dataset we (i) treat the risk–free rate as negligible, like $r_f \approx 0$. (ii) drop the constant $\sqrt{K}$, which only rescales all Sharpe values by the same factor. (iii) calculate the expectation and variation over $\mathcal{T}$. Throughout the paper we therefore report a raw Sharpe ratio of the form
\begin{equation}
  \mathcal{S}_\mathcal{T}
  \;=\;
  \frac{\mathbb{E}[R_t]}{\sqrt{\mathrm{Var}(R_t)}},
\end{equation}
and interpret it as a scale–free, risk–adjusted score rather than a directly tradable quantity.

Ideally, we would like the learned policy to achieve a good trade–off between cumulative return and Sharpe ratio. Conceptually, this can be formulated as a multi–objective optimization problem over the model parameters $\theta$,
\begin{equation*}
  \theta^* = \arg\max_{\theta} \;
  \Bigl(
    {ROI}_{\mathcal{T}}(\theta),
    \;\mathcal{S}_{\mathcal{T}}(\theta)
  \Bigr),
\end{equation*}
The exact loss components and their implementation details are given in Section~\ref{sec:loss-design}.

\subsection{Wavelet-based Signal Noise Decomposition}
At short horizons, financial returns exhibit low signal-to-noise ratios. Following wavelet denoising, we decompose the target log return into a slowly varying component and a rapidly fluctuating residual $\ell_{j^\star,t} = s_t + n_t$, 
where $s_t$ captures low-frequency structure and $n_t$ aggregates high-frequency, largely unpredictable variations. Over an interval $\mathcal{T}$, we define the effective SNR as
\begin{equation}
  \mathrm{SNR}_\mathcal{T}
  \;=\;
  \frac{\mathrm{Var}_{t \in \mathcal{T}}[s_t]}
       {\mathrm{Var}_{t \in \mathcal{T}}[n_t] + \varepsilon},
\end{equation}
with a small $\varepsilon>0$ for numerical stability.

In WaveLSFormer, a neural wavelet front-end performs a learnable time-frequency decomposition of the raw log-return sequence $\{\ell_{j^\star,t}\}$ into approximations of $\{s_t\}$ and $\{n_t\}$ using finite-impulse-response low-/high-pass filters. We learn filters that increase $\mathrm{SNR}$ relative to the raw series while producing features that benefit the downstream trading objective. To this end, we add frequency-domain regularizers that promote low-/high-frequency separation and encourage the filter pair to behave approximately as a tight frame; details are given in Section~\ref{sec:architecture}.

Given multivariate histories $\mathbf{X}_{t-1}$ and the next-step return $\ell_{j^\star,t}$, our goal is to learn a mapping $f_\theta$ that outputs trading signals $p_t$ whose induced positions $w_t$ maximize the performance of the risk-adjusted strategy, aided by the wavelet front-end that improves the effective SNR of the input. In the following chapters, $\ell_{j^\star,t}$ is simplified to $\ell_t$.

\subsection{Position Rule Under a Fixed Risk Budget}
\label{subsec:pos-rule}
\noindent
Eq.~\eqref{eq:tanh-position} bounds the trading signal, so a single trade never exceeds full long/short exposure. Yet different models can induce very different distributions of $w_t$. Some outputs saturate near $\pm 1$, while others concentrate around $0$. Using $w_t$ directly would therefore confound signal quality with aggressiveness, since strategies with larger average $|w_t|$ effectively take larger bets and thus realize larger gains and losses.

To normalize risk across models, we estimate a global scale on the validation set $\mathcal{D}_{\mathrm{val}}$ with length $T_{\mathrm{val}}=|\mathcal{D}_{\mathrm{val}}|$. We compute the mean absolute magnitude and rescale the signal using $\hat{w}_t$, so the validation-period average of $|\hat{w}_t|$ is close to one.
\begin{equation}
  s_{\mathrm{val}}
  = \frac{1}{T_{\mathrm{val}}}
    \sum_{t \in \mathcal{D}_{\mathrm{val}}} |w_t|, \qquad\hat{w}_t = \frac{w_t}{s_{\mathrm{val}}},
  \label{eq:actual-pos}
\end{equation}

To suppress tiny, noise-dominated trades, we apply a dead zone with threshold $\tau>0$ and set $\tau=0.01$ in all experiments. We then enforce an optional leverage cap and scale by a target average leverage $L>0$ (we use $L=1$), yielding the executable position in
\begin{equation}
  \tilde{w}_t =
  \begin{cases}
    0, & |\hat{w}_t| < \tau, \\
    \max(-L, \min(L, \hat{w}_t)), & \text{otherwise}.
  \end{cases}
  \label{eq:clamp-position}
\end{equation}

By this construction, the position is up to sparsification from the dead zone, so models are evaluated under a comparable risk budget. The scale $s_{\mathrm{val}}$ is estimated once on $\mathcal{D}_{\mathrm{val}}$ and kept fixed for testing and online simulation, ensuring a consistent and implementable mapping from network outputs to positions at a pre-specified risk level.
\begin{equation*}
  \frac{1}{T_{\mathrm{val}}}
  \sum_{t \in \mathcal{D}_{\mathrm{val}}} |\tilde{w}_t|
  \approx L,
\end{equation*}

\section{Data Construction Pipeline}
\label{sec:data-construction-pipeline}
Our model is designed to directly optimize a trading objective rather than minimize a point-wise forecasting error.
Therefore, we focus on economic performance, such as annualized return and Sharpe ratio, as the primary evaluation metrics, rather than MSE/MAE.

\subsection{Data Collection}
We construct an intraday U.S.\ equity dataset from two sources. We obtain price and volume data from the commercial API provider \textit{polygon.io}, and download one-hour OHLCV bars for each stock over roughly the past ten years, providing multi-year coverage at intraday resolution.

We obtain sector and industry classifications from \textit{StockAnalysis}\footnote{\url{https://stockanalysis.com/}}, which publishes curated U.S.\ stock lists by sector and industry. We extract symbols for each industry and use this stock-industry mapping to define the cross-sectional universe and the industry-group experiments in Section~\ref{sec:experiment}.

\subsection{Data Preprocessing}
We consider hourly bar-based intraday trading. Each stock is represented by OHLCV bars, and at the beginning of hour $t$ the model observes past bars and selects the position to hold over hour $t{+}1$, making within-bar price change a natural primitive. Because price levels vary widely across securities, we use scale-free inputs by converting prices to intraday returns: Eq.~\eqref{eq:price_return} computes the log percentage change of the close relative to the open, which stabilizes variance, partially symmetrizes gains and losses, and yields smoother dimensionless features for hourly modeling.

\subsection{Stock Selection}
\label{sec:stock-selection}
\begin{figure}
	\centering
	\includegraphics[width=\linewidth]{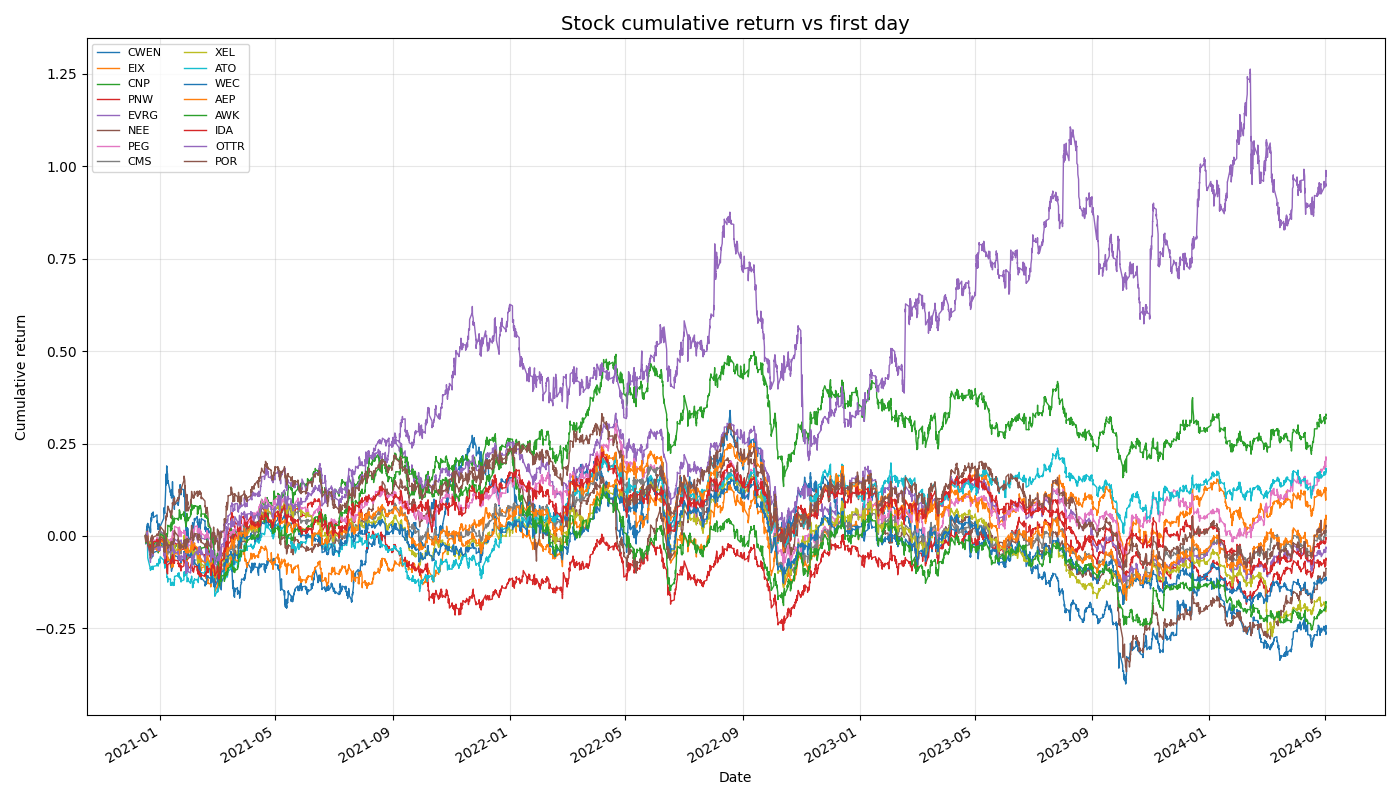}
	\caption{Stock codes selected under DWT.}
	\label{fig:DWT results}
\end{figure}

We build industry-level stock groups as multi-asset inputs. For each industry with a liquid U.S.-listed ETF, we collect constituent stocks traded on major U.S.\ exchanges and treat the ETF as an economic proxy for the sector. We quantify similarity in recent price dynamics by computing Dynamic Time Warping (DTW) distances between candidates' historical log-change series over a fixed look-back window, where DTW is robust to local time shifts. Using the training period, we set the empirical median (50th percentile) of DTW distances as a threshold and retain stocks whose distance to the industry reference is below this value, filtering idiosyncratic outliers and preserving coherent trend shapes as illustrated in Fig.~\ref{fig:DWT results}. We then apply non-parametric Granger causality tests to the DTW-selected set and remove stocks that show no predictive relation with the target asset.

\subsection{Financial Securities with Granger-type Causality}
\label{sec:granger}
We assume that same-industry stocks exhibit co-movements and lead-lag effects. We quantify such dependencies using nonparametric Granger causality~\cite{diks2006new}, where $x_t$ Granger-causes $y_t$ if adding lags of $x_t$ significantly improves predicting $y_t$ beyond using lags of $y_t$ alone.

\begin{figure}
  \centering
  \includegraphics[width=\linewidth]{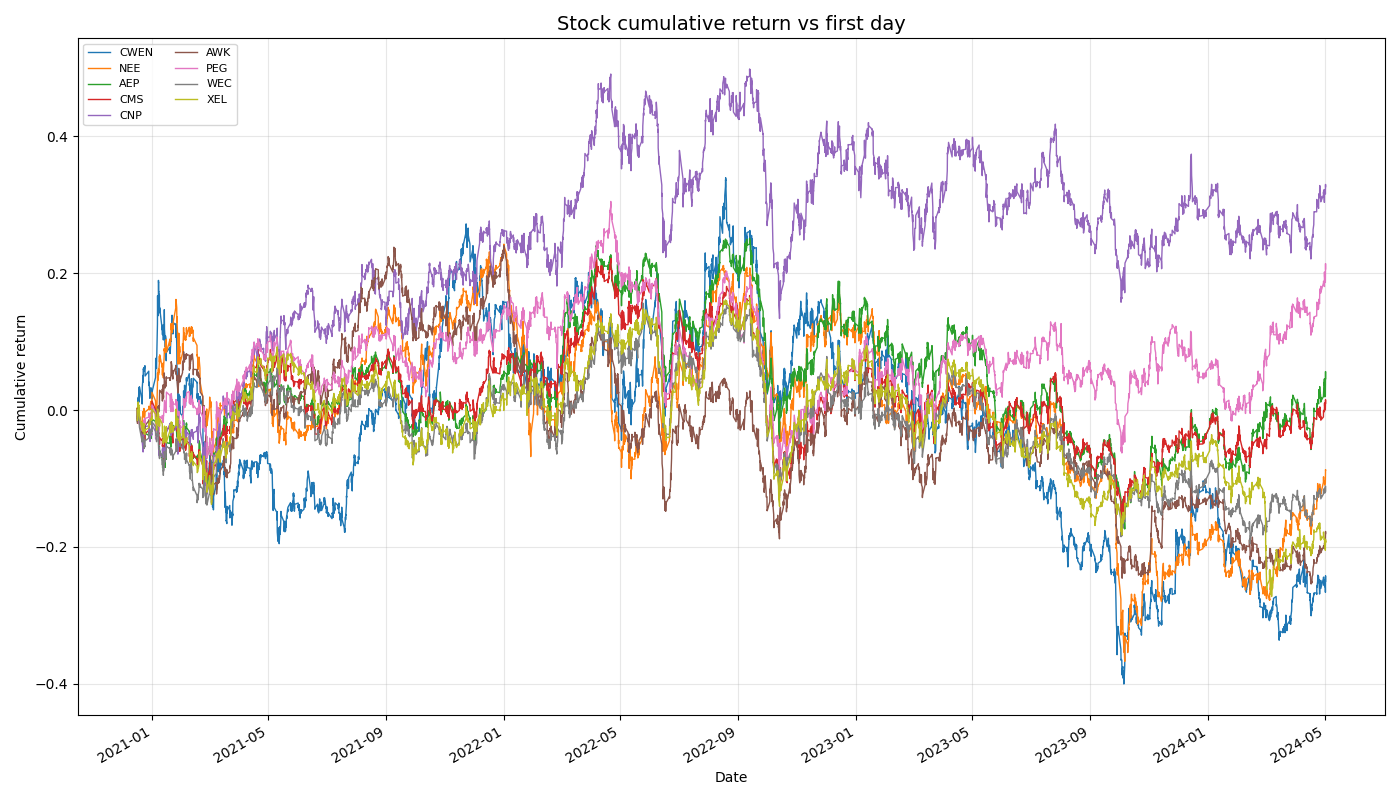}
  \caption{Stock selected after Granger causality filter in the renewable energy industry.}
  \label{fig:granger results}
\end{figure}

For each industry universe, we run pairwise tests on the log-change series of candidate securities to obtain a directional p-value matrix. We then apply Benjamini-Hochberg FDR control and retain securities whose adjusted p-values with respect to the target asset are below $0.05$ in at least one direction. The retained securities form the final multi-asset input set. Fig.~\ref{fig:granger results} illustrates the selected stocks in Renewable Energy, and Table~\ref{tab:logpct_corr} reports the full adjusted p-value matrix, which reveals the directed dependence structure within each industry universe.

\begin{table*}[!t]
\centering
\tiny
\renewcommand{\arraystretch}{1.0}
\setlength{\tabcolsep}{2pt}
\caption{BH-FDR-adjusted $p$-value matrix for pairwise tests on hourly log-percent-change series. Smaller values indicate stronger evidence of dependence. Bold entries denote significant pairs under FDR control ($p_{\text{adj}}<0.05$); bold tickers indicate the retained assets.}
\resizebox{\textwidth}{!}{%
\begin{tabular}{lcccccccccccccccc}
\hline
 & \cellcolor{gray!15}\textbf{AEP} & ATO & \cellcolor{gray!15}\textbf{AWK} & \cellcolor{gray!15}\textbf{CMS} & \cellcolor{gray!15}\textbf{CNP} & CWEN & EIX & EVRG & IDA & \cellcolor{gray!15}\textbf{NEE} & OTTR & \cellcolor{gray!15}\textbf{PEG} & PNW & POR & \cellcolor{gray!15}\textbf{WEC} & \cellcolor{gray!15}\textbf{XEL} \\
\hline
AEP  & 1 & 0.8107 & 0.7066 & 0.7824 & 1 & 0.3241 & 0.8397 & 0.9305 & 0.8041 & 0.8434 & 0.3984 & 0.4637 & 0.6562 & 0.9759 & 0.9506 & 0.8379 \\
ATO  & 0.3984 & 1 & 0.3984 & 0.3241 & 0.3984 & 0.6317 & 1 & 1 & 1 & 0.8041 & 0.3812 & 1 & 0.7737 & 0.9759 & 0.3812 & 0.5568 \\
AWK  & 1 & 0.7002 & 1 & 1 & 0.9759 & 0.0846 & 1 & 0.7393 & 0.8379 & 0.4262 & 0.456 & 0.8379 & 0.9305 & 0.8041 & 0.8379 & 0.6723 \\
CMS  & 0.0521 & 1 & 0.3716 & 1 & 0.9506 & 0.3984 & 0.8379 & 1 & 0.8379 & 0.3835 & 0.8379 & 1 & 0.8813 & 0.6592 & 0.3812 & 0.3487 \\
CNP  & 0.0505 & 0.3984 & 0.3241 & 0.4262 & 1 & 0.4262 & 0.9612 & 0.4838 & 0.7276 & 0.3984 & 0.3984 & 0.8107 & 0.3798 & 0.3984 & 0.3241 & 0.2143 \\
\cellcolor{gray!15}\textbf{CWEN} & \cellcolor{gray!15}\textbf{0.0124} & 0.0846 & \cellcolor{gray!15}\textbf{0.0326} & \cellcolor{gray!15}\textbf{0.0221} & \cellcolor{gray!15}\textbf{0.0297} & 1 & 0.0576 & 0.0648 & 0.1123 & \cellcolor{gray!15}\textbf{0.0001} & 0.7909 & \cellcolor{gray!15}\textbf{0.0326} & 0.3812 & 0.1703 & \cellcolor{gray!15}\textbf{0.0326} & \cellcolor{gray!15}\textbf{0.0345} \\
EIX  & 0.3812 & 0.8107 & 0.3487 & 0.5568 & 0.296 & 0.9955 & 1 & 1 & 0.8107 & 0.3984 & 0.7025 & 0.8379 & 0.3487 & 0.4262 & 0.4262 & 0.4701 \\
EVRG & 0.0297 & 0.2143 & 0.0149 & 0.0558 & 0.1048 & 0.8107 & 0.52 & 1 & 0.3716 & 0.1719 & 0.3241 & 0.3467 & 0.2443 & 0.3812 & 0.0505 & 0.0912 \\
IDA  & 0.4181 & 0.7737 & 0.4262 & 0.3812 & 0.1836 & 0.6062 & 0.8379 & 0.6235 & 1 & 0.3812 & 0.596 & 0.9759 & 0.3984 & 0.3342 & 0.3984 & 0.7409 \\
NEE  & 0.7909 & 0.3449 & 0.3812 & 0.4262 & 0.6396 & 0.8107 & 0.8731 & 0.503 & 0.2482 & 1 & 0.6235 & 0.8409 & 0.5043 & 0.3812 & 0.5805 & 0.4397 \\
OTTR & 1 & 0.6062 & 0.9759 & 0.942 & 0.8379 & 0.9506 & 0.8434 & 0.296 & 0.6235 & 1 & 1 & 0.6396 & 0.6592 & 0.9759 & 0.7096 & 0.8107 \\
PEG  & 0.0687 & 0.3812 & 0.0326 & 0.0846 & 0.4262 & 0.9421 & 0.5308 & 0.5006 & 0.604 & 0.0846 & 0.4106 & 1 & 0.3812 & 0.296 & 0.3984 & 0.296 \\
PNW  & 0.8379 & 0.9759 & 0.7409 & 1 & 1 & 0.5985 & 0.7393 & 0.9759 & 1 & 0.3241 & 1 & 1 & 1 & 0.9759 & 0.9305 & 0.9633 \\
POR  & 0.3984 & 0.8813 & 0.3161 & 0.3984 & 0.4713 & 0.9791 & 0.9759 & 0.8321 & 0.9759 & 0.2213 & 0.6988 & 0.8107 & 0.7224 & 1 & 0.3812 & 0.8321 \\
WEC  & 0.531 & 1 & 0.4262 & 0.4262 & 1 & 0.3487 & 0.7909 & 0.8379 & 0.7409 & 0.4838 & 0.8379 & 0.8713 & 0.9421 & 0.9759 & 1 & 0.2443 \\
XEL  & 1 & 0.9305 & 0.9573 & 0.9759 & 0.9123 & 0.3812 & 0.4397 & 0.8813 & 0.9305 & 0.8379 & 0.8321 & 0.3812 & 0.9791 & 0.9759 & 1 & 1 \\
\hline
\end{tabular}}
\label{tab:logpct_corr}
\end{table*}

\section{Architecture}
\label{sec:architecture}
WaveLSFormer combines a learnable wavelet front-end with a standard Transformer. It decomposes prices into low-frequency trends and high-frequency residuals, fuses them, and feeds a Transformer encoder for position prediction. We therefore detail the front-end—filter parameterization, regularization, and cross-frequency integration that injects high-frequency cues into the low-frequency path.

\subsection{Neural Wavelet Filters}
As widely known, financial data contains a low info-noise ratio that causes the training process to always be misled by high-frequency noise gradient. To deal with this issue, we implement neural wavelet blocks as roles of low-pass and high-pass filters to re-balance the proportion of input info and noise. 

\subsubsection{FIR Convolution Kernels and Classical Discrete–time Filters}
\label{sec:FIR}
In our neural wavelet front-end, each learnable filter is implemented as a one-dimensional finite impulse response (FIR) convolution kernel. Let $x[n]$ denote a discrete-time input signal and let $\boldsymbol{\theta} = (\theta_0,\dots,\theta_{L-1})$ be the convolution kernel of length $L$. A standard \texttt{conv1d} layer with stride~1 and no dilation computes
\begin{equation}
  y[n] = \sum_{k=0}^{L-1} \theta_k\, x[n-k],
  \label{eq:fir-conv}
\end{equation}
which is exactly the input and output relation of a linear time-invariant (LTI) discrete-time filter.

For an LTI system, the sequence $h[n]$ is defined as the response to a discrete-time unit impulse
\begin{equation*}
    \delta[n] =
  \begin{cases}
    1, & n = 0,\\
    0, & n \neq 0
  \end{cases}
\end{equation*}

Substituting $x[n] = \delta[n]$ into Eq.~\eqref{eq:fir-conv} yields
\begin{equation*}
  y[n]
  = \sum_{k=0}^{L-1} \theta_k\, \delta[n-k]
  = \theta_n,
\end{equation*}
for $n=0,\dots,L-1$ and $y[n]=0$ otherwise. Hence the impulse response of the filter is $h[n] = \theta_n$, so the FIR convolution kernel and the time-domain impulse response are the same object that the trainable parameters $\{\theta_k\}$ directly define the filter's behavior in the time domain.

The discrete-time frequency response of an LTI filter is defined as the complex gain applied to a complex exponential input $x[n] = e^{\mathrm{j}\omega n}$.  Using Eq.~\eqref{eq:fir-conv} we obtain
\begin{align*}
  y[n]
  &= \sum_{k=0}^{L-1} \theta_k\, e^{\mathrm{j}\omega (n-k)} 
  = e^{\mathrm{j}\omega n}
     \sum_{k=0}^{L-1} \theta_k\, e^{-\mathrm{j}\omega k}.
\end{align*}

Therefore, $e^{\mathrm{j}\omega n}$ is an eigenfunction of the system and the corresponding eigenvalue
\begin{equation}
  H(e^{\mathrm{j}\omega})
  = \sum_{k=0}^{L-1} \theta_k\, e^{-\mathrm{j}\omega k}
  = \sum_{k=0}^{L-1} h[k]\, e^{-\mathrm{j}\omega k}
  \label{eq:freq-response}
\end{equation}
is the frequency response. Eq.~\eqref{eq:freq-response} is exactly the discrete-time Fourier transform (DTFT) of the impulse response $h[n]$. Therefore, for an FIR filter, the frequency response is simply the Fourier transform of the convolution kernel.  This provides a direct bridge between the learnable parameters in the neural network and the classical time–frequency interpretation of digital filters.

\subsubsection{Differentiable Spectral Regularization via rFFT}
Let $\boldsymbol{\theta}\in\mathbb{R}^{L}$ be a real-valued FIR kernel and $\hat{\boldsymbol{\theta}}=\mathcal{F}(\boldsymbol{\theta})$ its DFT. 
In matrix form, $\hat{\boldsymbol{\theta}}=\mathbf{F}\boldsymbol{\theta}$ with a fixed $\mathbf{F}\in\mathbb{C}^{L\times L}$. Hence $\mathcal{F}$ is linear and differentiable everywhere, with constant Jacobian $\mathbf{F}$. 
In practice we use rFFT, which is an optimized implementation of the same linear map specialized to real inputs.

This enables differentiable frequency-domain regularization. A generic spectral regularizer is
\begin{equation}
  \mathcal{L}_{spec}(\boldsymbol{\theta})
  = \sum_{k} \rho\!\left(|\hat{\theta}_k|\right),
  \label{eq:spectral-regularizer}
\end{equation}
where $\rho(\cdot)$ is differentiable. By the chain rule, the gradient of spectral regularizer with respect to the time-domain kernel is
\begin{equation*}
  \frac{\partial \mathcal{L}_{spec}}{\partial \boldsymbol{\theta}}
  = \mathbf{F}^\mathsf{H}\,
    \frac{\partial \mathcal{L}_{spec}}{\partial \hat{\boldsymbol{\theta}}},
\end{equation*}
where $\mathbf{F}^\mathsf{H}$ denotes the Hermitian transpose of $\mathbf{F}$, i.e.\ the inverse DFT up to a constant scaling factor. Equivalently, back-propagation through rFFT amounts to applying an inverse FFT to the frequency-domain gradient. Therefore, adding Eq.~\eqref{eq:spectral-regularizer} on rFFT-transformed learnable kernels preserves differentiability and allows end-to-end training with standard optimizers.

To enforce complementary low/high-frequency coverage, we penalize high-frequency energy in the low-pass filter and low-frequency energy in the high-pass filter, add an overlap term to sharpen separation, and use Parseval and shape losses to constrain energy and frequency-response form; details are in Section~\ref{sec:wavelet_loss}.

\subsubsection{Low-guided High-frequency Injection}
\label{subsec:gamma-in-lowguided-injection}
To fuse low-frequency and high-frequency information while maintaining training stability, we introduce a low-guided high-frequency injection (LGHI) module.
Let $L \in \mathbb{R}^{T\times d}$ denote the low-frequency representations and $H \in \mathbb{R}^{T\times d}$ the corresponding high-frequency representations.
Different from standard cross-attention, where $Q$ comes from one sequence and $K,V$ come from the other, we compute the attention map \emph{only} from the low-frequency branch and use it to inject high-frequency values.
Concretely, we define
\begin{equation*}
\begin{split}
&A(L) = \mathrm{softmax}\!\left(\frac{(L W_Q)(L W_K)^{\top}}{\sqrt{d_k}}\right),\\
&Z(L,H) = A(L)\,(H W_V) W_O,
\end{split}
\end{equation*}
where $W_Q, W_K \in \mathbb{R}^{d\times d_k}$ and $W_V \in \mathbb{R}^{d\times d_v}$ are projection matrices and
$W_O$ is the output projection.
The final fused representation is given by a gated residual injection:
\begin{equation}
  Y = L + \beta\, Z(L,H),
  \qquad
  \beta = \sigma(\gamma),
  \label{eq:gated_lowguided_injection}
\end{equation}
where $\gamma$ is a learnable scalar controlling the overall scale of the injected high-frequency residual.
This design encourages the model to preserve the stable low-frequency backbone while selectively incorporating high-frequency cues, because the attention weights are determined by $L$ and are therefore less sensitive to noisy high-frequency fluctuations.

\section{Training Design}
\label{sec:training-design}
Training choices, such as loss, optimizer, and learning-rate schedule, strongly influence trading outcomes. Because convenient objectives may misalign with deployment metrics, such as return, Sharpe and draw-down, low training loss can still generalize poorly. We therefore present: (i) a differentiable surrogate objective (soft labels from future log returns + Sharpe regularization), (ii) a stabilized optimization setup for heavy-tailed financial noise, and (iii) validation/checkpoint selection based on trading performance.

\subsection{Loss Function Design}
\label{sec:loss-design}
In stock trading, the ultimate objective is to maximize realized profit rather than to minimize point-wise prediction error. Therefore, the training loss should be a simple, differentiable surrogate that aligns with return optimization and provides stable gradients. In generic time-series forecasting, models are typically trained to predict the next-step target value, where MAE or MSE are standard choices. However, in financial settings, minimizing MAE/MSE on prices or returns is often weakly correlated with trading performance and thus is not well suited for learning profit-seeking decisions.

\subsubsection{Drawbacks of Regression Losses}
Many financial time-series models train neural networks as return forecasters by minimizing point-wise regression losses such as MSE or MAE. Given a future log return $\ell_t$ as the ground truth, they train $f_{\theta}(\mathbf{X}_{t-1})$ via
\begin{equation}
    \mathcal{L}_{reg}
    = \mathbb{E}\big[ loss(f_{\theta}(\mathbf{X}_{t-1}),\ell_t) \big],
    \qquad
    loss \in \{\text{MSE},\ \text{MAE}\}.
    \label{eq:regression-loss}
\end{equation}

While statistically convenient, $\mathcal{L}_{reg}$ is misaligned with trading: performance is driven by the position $\tilde{w}_t$ and the strategy return profit $R_t=\tilde{w}_t\,f_{\theta}(\mathbf{X}_{t-1})$, not by the numeric forecast error of $f_{\theta}(\mathbf{X}_{t-1})$. Consequently, lower regression error does not necessarily translate to higher ROI/Sharpe. The mismatch is exacerbated by the asymmetric economic cost as a wrong sign under high exposure is far more damaging than a small magnitude error on a correct trade, whereas MSE/MAE penalize deviations symmetrically.

Moreover, under noisy and non-stationary returns, minimizing $\mathcal{L}_{reg}$ mainly encourages approximating $\mathbb{E}[f_{\theta}(\mathbf{X}_{t-1})]$, which offers no guarantee of improved $R_t$ or risk-adjusted performance. Therefore, we do not treat the network as a pure forecaster. Instead, we interpret its output as a continuous trading position $\tilde{w}_t\in[-1,1]$ (flat if $\tilde{w}_t\in[-0.01,0.01]$) and optimize a trading-aligned objective 
\begin{equation*}
    \mathcal{L}_{trade}
    = -\,\mathbb{E}\big[\tilde{w}_t \ell_t\big]
      + \lambda\,\Omega(\tilde{w}_t),
\end{equation*}
where $\Omega(\cdot)$ regularizes leverage, turnover, or excessive position variability.

\subsubsection{Stock Direction and Tanh Position Loss}
The \emph{Stock Direction} loss treats the sign of the model output as a long/short signal: go long if $p_t>0$,  short if $p_t<0$, and optionally trade only when $|p_t|>\tau$ to filter low-confidence positions. The resulting position in Eq.~\eqref{eq:roi} is defined in Eq.~\eqref{eq:tanh-position}. Following Eq.~\eqref{eq:price_return}, let $r_t$ and $\ell_t$ denote the simple and log returns.

To turn multiplicative ROI into additive form, we use the small-return approximation. For $|r_t|<0.1$,
\[
\log(1+r_t)=r_t-\frac{r_t^2}{2}+O(r_t^3),
\]
so $\ell_t\approx r_t$ with leading error $r_t^2/2\le 0.005$ when $|r_t|\le 0.1$. Hence we approximate
\begin{equation*}
\mathrm{ROI}\approx \exp\Bigl(\sum_t \mathrm{sign}(p_t)\,\ell_t\Bigr)-1,
\end{equation*}
and define the corresponding trade loss as
\begin{equation*}
\mathcal{L}_{trade} = -\exp\Bigl(\sum_t \mathrm{sign}(p_t)\,\ell_t\Bigr).
\end{equation*}

The \emph{Tanh position} loss replaces the all-in direction decision by a continuous position size on the target asset using $\tanh(p_t)\in[-1,1]$ as
\begin{equation}
\mathcal{L}_{trade} = -\exp\Bigl(\sum_t \tanh(p_t)\,\ell_t\Bigr).
\label{eq:tanh_loss}
\end{equation}

\subsubsection{Sigmoid Position Loss}
Direct position optimization with the Stock Tanh loss often suffers from gradient saturation: as outputs approach the bounds, $\tanh'(\cdot)$ becomes small and updates vanish. We also avoid $\log(\tanh(\cdot))$ objectives, whose logit gradient
\begin{equation*}
\frac{\mathrm{d}}{\mathrm{d}p_t}\log(\tanh(p_t))
= \frac{1-\tanh^2(p_t)}{\tanh(p_t)},
\end{equation*}
is ill-conditioned near the origin and still decays as $|p_t|$ grows. Instead, we treat $p_t$ as the logit of a Bernoulli variable indicating willingness to go long and perform the signed position mapping only at inference.

Ignoring transaction costs and position constraints, the ROI-optimal policy is fully long for $r_t\ge 0$ and fully short otherwise, so we cast trading as a binary classification problem:
\begin{equation*}
    y_t =
    \begin{cases}
    1, & r_t \ge 0,\\
    0, & r_t < 0.
    \end{cases}
\end{equation*}
We train in probability space with a weighted logistic loss, so gradients depend on the confidence gap between $\sigma(p_t)$ and $y_t$ and remain informative; at inference, we map $\sigma(p_t)$ to a smooth exposure in $[-1,1]$ via Eq.~\eqref{eq:tanh-position}.

To emphasize economically significant moves, we weight each sample by the absolute log return $|\ell_t|$, i.e., $+100\%$ and $-50\%$ have equal $|\ell_t|$. The resulting Stock Sigmoid loss is
\begin{equation}
\mathcal{L}_{trade}(t) =
\begin{cases}
- \log\big(\sigma(p_t)\big)\,|\ell_t|, & r_t \ge 0, \\[4pt]
- \log\big(1 - \sigma(p_t)\big)\,|\ell_t|, & r_t < 0,
\end{cases}
\end{equation}
which mitigates saturation in Stock Tanh while aligning optimization with the sign and magnitude of future returns.

\subsubsection{Soft-Label Position Loss}
Although the Stock Sigmoid loss mitigates the saturation of Stock Tanh, it still has two drawbacks: gradients are dominated by rare extreme moves, and hard binary targets remain overly sharp when returns are near zero, forcing $P_t=\sigma(p_t)$ toward $0$ or $1$ despite highly ambiguous signals, which can induce unstable updates.

We therefore replace hard labels with probabilistic soft targets that reflect directional confidence $y_t=\sigma(k\,\ell_t)$, 
where $k>0$ controls how quickly $y_t$ departs from $0.5$ as $|\ell_t|$ increases. We choose $k$ by calibration (e.g., enforcing $\sigma(k\,\ell_t^{(+5\%)})\approx 0.9$), which yields $k\approx 45$ in our setting; we fix $k=45$ for all experiments across industries for reproducibility. Given $y_t$ and $P_t$, we use the soft-label cross-entropy
\begin{equation}
    \mathcal{L}_{trade}(t)
    = -y_t \log P_t - (1-y_t)\log(1-P_t).
\end{equation}

At inference time, we map the model output to a continuous position in $[-1,1]$ using Eq.~\eqref{eq:clamp-position}, where the sign encodes long, short and flat. The magnitude controls position size. Overall, soft labels provide smoother gradients, reduce over-emphasis on extreme returns, and align the learned signal with probabilistic directional confidence.

\textit{Remark.} Subtracting the Bernoulli entropy of $y_t$ turns the above loss into a KL divergence and shifts the optimum to zero, but it does not affect gradients; we therefore use the standard form for simplicity.

\subsubsection{Overfitting Penalty}
Directly optimizing a profit-oriented objective can encourage unrealistic, overfit strategies with extremely large training returns. We therefore impose a soft upper bound on batch-level ROI.

For a mini-batch $\mathcal{B}$ spanning $H_{\mathcal{B}}$ time steps, the batch log return and ROI are
\begin{equation*}
    L_{\mathcal{B}}=\sum_{t\in\mathcal{B}}\ell_t,
    \qquad
    R_{\mathcal{B}}=\exp(L_{\mathcal{B}})-1.
\end{equation*}

We cap the annualized ROI at $R_{\mathrm{ann}}^{\max}=1.0$ and treat it as a tunable risk-control parameter. With hourly data, $252$ trading days/year and $6.5$ hours/day, we have $H_{\mathrm{year}}=1638$. Assuming constant per-hour log return, the implied maximum batch log return is $L_{\mathcal{B}}^{\max}$, which yields the batch ROI threshold $T_{\mathcal{B}}$.
\begin{equation*}
    L_{\mathcal{B}}^{\max} = \log(1+R_{\mathrm{ann}}^{\max})\,\frac{H_{\mathcal{B}}}{H_{\mathrm{year}}},
    \qquad
    T_{\mathcal{B}} = \exp(L_{\mathcal{B}}^{\max})-1
    = (1+R_{\mathrm{ann}}^{\max})^{H_{\mathcal{B}}/H_{\mathrm{year}}}-1
\end{equation*}

We then penalize excess ROI using a one-sided quadratic hinge, where $\lambda_{\mathrm{roi}}$ controls the penalty strength.
\begin{equation}
    \mathcal{L}_{penalty}
    =
    \lambda_{\mathrm{roi}}
    \bigl[\max(R_{\mathcal{B}}-T_{\mathcal{B}},\,0)\bigr]^2,
\end{equation}

\subsubsection{Sharpe Regularizer}
To account for risk, we augment the trading loss with a Sharpe-ratio regularizer. Let $R_p$ be the per-step strategy return. The classical Sharpe is
\begin{equation*}
    S^\ast = \frac{\mathbb{E}[R_p]-r_f}{\sigma(R_p-r_f)},
\end{equation*}
where $r_f$ is the risk-free rate. We omit subtle $r_f$ and use $S = \frac{\mathbb{E}[R_p]}{\sigma(R_p)}$ with $\mathbb{E}[\cdot]$ and $\sigma(\cdot)$ computed over a mini-batch. This is justified by our short horizon, so that $r_f$ is small relative to $R_p$, and by $r_f$ being approximately constant w.r.t.\ model parameters, hence having negligible impact on gradient directions. For simplicity, we ignore transaction costs and assume $R_p$ has finite second moments.

Instead of maximizing $S$ directly, we use an exponentially decaying penalty with a capped Sharpe level:
\begin{equation}
    \mathcal{L}_{sharpe}
    = \exp\!\left(
        -\alpha \cdot 
        \min\!\Big(\frac{3}{\sqrt{K}},\; \frac{\mathbb{E}[R_p]}{\sigma(R_p)+\varepsilon}\Big)
      \right),
\end{equation}
where $K$ is the annualization factor in Eq.~\eqref{eq:sharpe_def}, $\alpha>0$ controls strength, and $\varepsilon$ stabilizes training. The cap $3/\sqrt{K}$ prevents unrealistically large Sharpe values from dominating; once exceeded, this term provides no additional gradient. The exponential form smooths optimization and avoids Sharpe-driven gradients overwhelming the primary loss.

For $\mathcal{L}_{sharpe}$ we compute $R_p$ without position clamping as defined in Eq.~\eqref{eq:clamp-position} to preserve gradients, while all reported performance uses clamped positions.

\subsubsection{Neural Wavelet Loss}
\label{sec:wavelet_loss}
Our low-/high-pass filters are implemented as 1D FIR convolutions. As discussed in Section~\ref{sec:FIR}, the convolution weights $W$ correspond to the impulse responses, whose discrete Fourier transform gives the frequency responses. Therefore, the rFFT of the filter parameters directly yields the spectrum on the grid
\begin{equation*}
    \frac{k}{n_{\mathrm{fft}}}, \qquad k\in \Bigl[0,1,\ldots,\frac{n_{\mathrm{fft}}}{2}+1\Bigr].
\end{equation*}

With hourly data, we set $n_{\mathrm{fft}}=81$ to cover month-level frequencies. We regularize the learnable filter bank in the frequency domain to encourage clear low-/high-pass separation and avoid degenerate solutions. Let $\lvert G_{\text{low}}\rvert^2:=\sum_{\omega}\lvert G_{\text{low}}(\omega)\rvert^2$ and $\lvert G_{\text{high}}\rvert^2:=\sum_{\omega}\lvert G_{\text{high}}(\omega)\rvert^2$ be the total spectral energies on a discrete grid $\omega\in[0,\pi]$. We penalize out-of-band energy using a frequency-weighted power term with $p=2$ in all experiments:
\begin{equation*}
\begingroup
\setlength{\jot}{2pt}
\begin{aligned}
&\mathcal{L}_{low} = \sum_\omega (\omega/\pi)^{p}\,\lvert G_{\text{low}}(\omega)\rvert^2, \qquad
\mathcal{L}_{high} = \sum_\omega (1-\omega/\pi)^{p}\,\lvert G_{\text{high}}(\omega)\rvert^2,\\
&\mathcal{L}_{overlap} = \lvert G_{\text{low}}\rvert^{2}\cdot \lvert G_{\text{high}}\rvert^{2}, \qquad
\mathcal{L}_{parseval} = \big(\lvert G_{\text{low}}\rvert^2+\lvert G_{\text{high}}\rvert^2-2\big)^{2}.
\end{aligned}
\endgroup
\end{equation*}

To balance the two branches, we further impose an energy-ratio hinge loss. Define $\rho$ as following and penalize $\rho$ outside $[\rho_{\min},\rho_{\max}]$
\begin{equation}
\rho = \frac{\lvert G_{\text{high}}\rvert^2}{\lvert G_{\text{low}}\rvert^2 + \varepsilon},\qquad
\mathcal{L}_{ratio} =
  \max\bigl( \rho - \rho_{\max},\, 0 \bigr)
  +
  \max\bigl( \rho_{\min} - \rho,\, 0 \bigr).
\label{eq:energy-ratio-and-loss}
\end{equation}

Finally, the neutral wavelet loss is a weighted sum
\begin{equation*}
\mathcal{L}_{wavelet}
= \lambda_\mathrm{spec} \bigl(\mathcal{L}_{low} + \mathcal{L}_{high}\bigr)
+ \mathcal{L}_{overlap}
+ \mathcal{L}_{parseval}
+ \mathcal{L}_{ratio},
\end{equation*}

\subsection{Validation Metric and Model Selection}
\label{sec:validation metric and model selection}
After the discussion above, the overall training objective is
\begin{equation}
   \mathcal{L}_{train}
   = \mathcal{L}_{trade} + \mathcal{L}_{penalty} + \mathcal{L}_{sharpe} + \mathcal{L}_{wavelet}.
   \label{eq:loss}
\end{equation}

We train the network by minimizing the loss in Eq.~\eqref{eq:loss}, but our goal is to maximize out-of-sample trading performance rather than validation cross-entropy. Using the probability-to-position rule in Eq.~\eqref{eq:clamp-position}, the cumulative strategy return on the validation set is
\begin{equation*}
    R_{\text{val}}
    = \sum_{t \in \mathcal{D}_{\text{val}}} \tilde{w}_t r_t.
\end{equation*}
In the risk-adjusted setting, we also compute a Sharpe-like ratio from the sequence $\{\tilde{w}_t r_t\}$ as an auxiliary validation metric.

Crucially, the validation loss $\mathcal{L}_{val}$ and $R_{\text{val}}$ are not necessarily monotonic. The threshold-based decision rule in Eq.~\eqref{eq:clamp-position} can flip actions near boundaries with minimal change in $\mathcal{L}_{val}$ yet a large change in $R_{\text{val}}$. Moreover, cross-entropy weights samples roughly uniformly, whereas $R_{\text{val}}$ is dominated by a small number of large moves: improving calibration on many small-return samples may reduce $\mathcal{L}_{val}$ with little impact on $R_{\text{val}}$, while better capturing large profitable moves can slightly worsen $\mathcal{L}_{val}$ but substantially improve $R_{\text{val}}$.

Therefore, we use soft-label cross-entropy as a differentiable surrogate for optimization, but perform early stopping and hyperparameter selection solely based on $R_{\text{val}}$. In practice, we select the checkpoint with the highest validation ROI, using $\mathcal{L}_{val}$ only as a diagnostic for optimization stability.

\subsubsection{Regression Loss vs. Soft-label Loss}
\label{sec:reg-vs-soft}
Motivated by the above discussion, we replace regression with a soft-label classification objective that focuses on return direction and economically relevant magnitude, while treating the output as a proxy for position.

With regression, the model minimizes Eq.~\eqref{eq:regression-loss} to predict the future log return. Trading then relies on a separate hand-designed mapping $h:\mathbb{R}\!\to\![\tilde{w}_{\min},\tilde{w}_{\max}]$, such as the clamping rule in Eq.~\eqref{eq:clamp-position}, and sets $\tilde{w}_t=h(\ell_t)$. As a result, $L_{\text{reg}}$ optimizes only return accuracy, whereas ROI/Sharpe depends on the untrained composite decision rule and can differ substantially even when $L_{\text{reg}}$ is similar.

To remove this layer, we form a soft target using a scaled sigmoid $y_t=\sigma(k\,\hat{\ell}_t)$, where $k>0$ controls saturation. The network outputs a logit $p_t$ with $P_t=\sigma(p_t)$ and is trained with soft-label cross-entropy. At inference, we convert $p_t$ to an executable position using the same squashing and risk-budget mapping described in Section~\ref{sec:problem-formulation}, so the optimized scalar becomes the trading signal up to a fixed monotone transform.

Soft labels offer three advantages: they penalize sign errors on large moves more than small deviations, learn a decision signal rather than a return forecast, and integrate naturally with the Sharpe-oriented regularizer in Section~\ref{sec:loss-design}.

Table~\ref{tab:reg-vs-soft-semi} shows Renewable Energy results with identical architecture and optimization. We disable the Sharpe regularizer and only change the supervised loss (MSE, MAE, or Soft-Label). The soft-label objective yields higher ROI and Sharpe than regression despite comparable error on $\ell_t$, supporting soft-label training for trading signals.

\begin{table}[!h]
  \centering
  \setlength{\tabcolsep}{8pt}
  \caption{Comparison of regression losses and soft-label loss on Renewable Energy with 10 seeds. All models share the same WaveLSFormer architecture and optimization setup, the Sharpe regularizer is disabled for all rows to isolate the effect of the supervised loss.}
  \label{tab:reg-vs-soft-semi}
  \renewcommand{\arraystretch}{1.0}
  \begin{tabular}{lcc}
    \toprule
    Loss                & ROI (test) & Sharpe (test) \\
    \midrule
    MSE                  &  $0.078 \pm 0.039$ & $0.586 \pm 0.275$ \\
    MAE                  &  $0.125 \pm 0.061$ & $0.899 \pm 0.355$ \\
    Soft-label           &  $0.377 \pm 0.045$ & $1.943 \pm 0.311$ \\
    \bottomrule
  \end{tabular}
\end{table}

\subsubsection{Tanh Loss vs. Soft-label Loss}
We observed that optimizing Transformers for U.S.\ equity trading is highly sensitive to the loss design. When using the tanh-based objective in Eq.~\eqref{eq:tanh_loss}, training often stagnates because the tanh nonlinearity saturates, producing near-zero gradients and effectively deadlocking the optimizer. We diagnosed this behavior by monitoring the $L_2$ norm of parameter gradients, which frequently collapsed toward zero, and found that an overly small initial learning rate further increases the likelihood of getting trapped in poor local optima.
However, the soft-label loss overcomes this drawback and produces a stable training process.
\begin{figure}
    \centering
    \begin{subfigure}{0.8\linewidth}
        \centering
        \includegraphics[width=\linewidth]{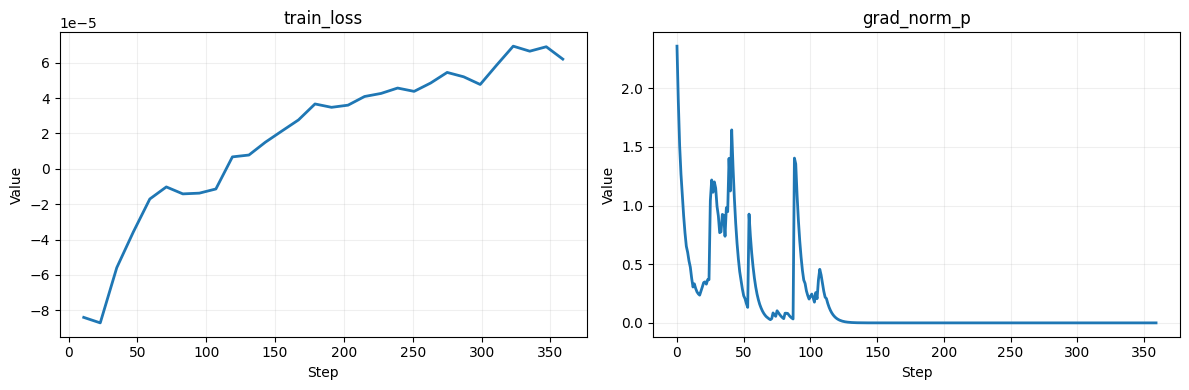}
	\caption{Tanh loss function leads to gradient vanishing.}
	\label{fig:gradient-vanish}
    \end{subfigure}
    \vspace{0.0 cm}
    \begin{subfigure}{\linewidth}
        \centering
        \includegraphics[width=0.8\linewidth]{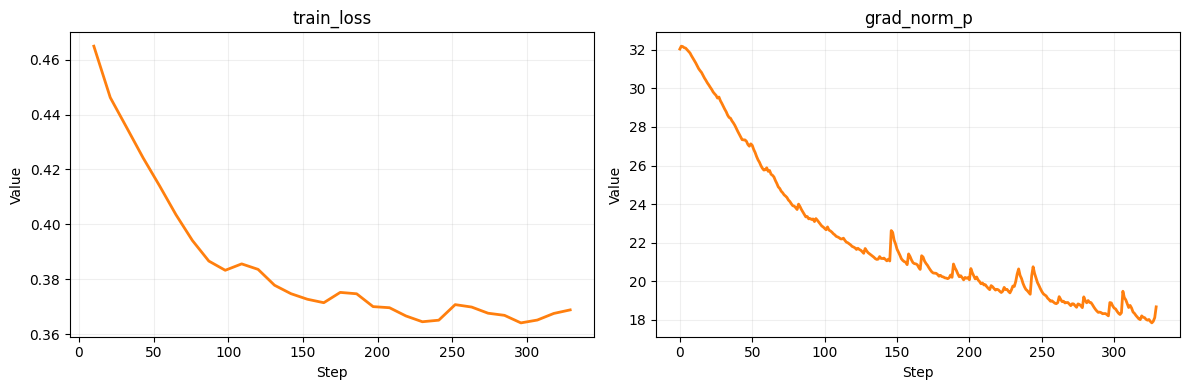}
    \caption{Soft-label loss function has stable and continuous gradients.}
    \label{fig:soft-label-gradient}
    \end{subfigure}
    \caption{Loss curve and gradient of tanh and soft-label loss functions}
\end{figure}

\section{Experiment}
\label{sec:experiment}
\subsection{Experiment Setup}
\label{subsec:experiment-setup}
\subsubsection{Data Split and Leakage Prevention}
We use five years of hourly U.S.\ equity data from 29.10.2020 to 29.10.2025 and adopt a temporal $0.7:0.1:0.2$ split for training, validation, and test, corresponding to 5292, 756, and 1512 time steps.
Due to limited computational resources, hyperparameters are tuned manually.

To prevent leakage, all preprocessing is computed on the training period only, including industry selection by an ARR threshold, DTW-based similarity filtering, and Granger-causality filtering with BH-FDR correction.
All models use identical per-industry datasets produced by this training-only pipeline.
For efficiency, Transformer-based variants use ProbSparse attention with distillation.

\subsubsection{Implementation and Model Selection}
After forming industry-level asset groups, we compare WaveLSFormer against two baselines, LSTM and MLP.
Unless otherwise stated, WaveLSFormer uses $d_{\text{model}}=512$, $d_{\text{ff}}=1024$, 6 encoder layers with $n_{\text{heads}}=128$, input length 96, and a 128-dimensional time2vec temporal embedding.

Each model is trained with ten random seeds for 80 epochs using batch size 256 and learning rate $10^{-5}$.
We adopt a two-phase schedule for the neural wavelet front-end: the first 30 epochs stabilize the learnable filters under spectral regularization, and the remaining epochs select checkpoints by validation trading metrics.
For each seed, we choose the checkpoint with the highest validation ROI in the second phase and evaluate it on the test set.
We report mean$\pm$std across the ten runs, and all results are computed on the held-out test set.

\subsection{Model Performance}
\label{subsec:model-performance}
Among 14 candidate industries in Table~\ref{tab:industry_filter}, we compute the industry-level annualized rate of return (ARR) of an intraday trading strategy using the training period only and retain industries with ARR $\ge 10\%$. The selected set is then fixed for all subsequent experiments to avoid leakage and to focus on sectors where intraday trading is potentially profitable and sufficiently liquid. This procedure selects six industries, and we focus on Renewable Energy and Retail Consumer Goods for detailed analysis, due to their having the lowest ARR.

\begin{table}[!t]
\centering
\setlength{\tabcolsep}{10pt}
\caption{Industry-level annualized rate of return (ARR) used for sector selection. Industries with ARR $\geq 10\%$ are retained for subsequent experiments.}
\label{tab:industry_filter}
\renewcommand{\arraystretch}{1}
\begin{tabular}{llc}
\toprule
Industry & ARR (\%) & Selected \\
\midrule
Biotechnology                     & 33.98505   & Yes \\
Regional Banks                    & 5.706405   & No  \\
Engineering Construction          & 8.208481   & No  \\
Electronic Components             & 8.899873   & No  \\
Information Technology Services   & 8.718001   & No  \\
Medical Devices                   & 12.48685   & Yes \\
Semiconductors                    & 14.11355   & Yes \\
Software Application              & 7.635932   & No  \\
Specialty Industrial Machinery    & 4.412272   & No  \\
Utilities Electric                & 4.132072   & No  \\
Real Estate REITs                 & 2.451229   & No  \\
\textbf{Renewable Energy}         & 10.14711   & Yes \\
Life Insurance                    & 12.77445   & Yes \\
\textbf{Retail Consumer Goods}    & 11.83257   & Yes \\
\bottomrule
\end{tabular}
\end{table}

For each selected industry, the constituent stocks are fed into the model following the pipeline described above. During training, we choose the model checkpoint that achieves the highest return on the validation set. For evaluation, we report trading performance under three execution modes: (i) Long \& Short, (ii) Long Only, and (iii) Short Only. These three modes jointly reflect the model's ability to exploit both upward and downward price movements. Additional per-industry trading performance curves are provided in the supplementary material, Figs.~S1-S4.

\begin{figure}
    \centering
    \includegraphics[width=\linewidth]{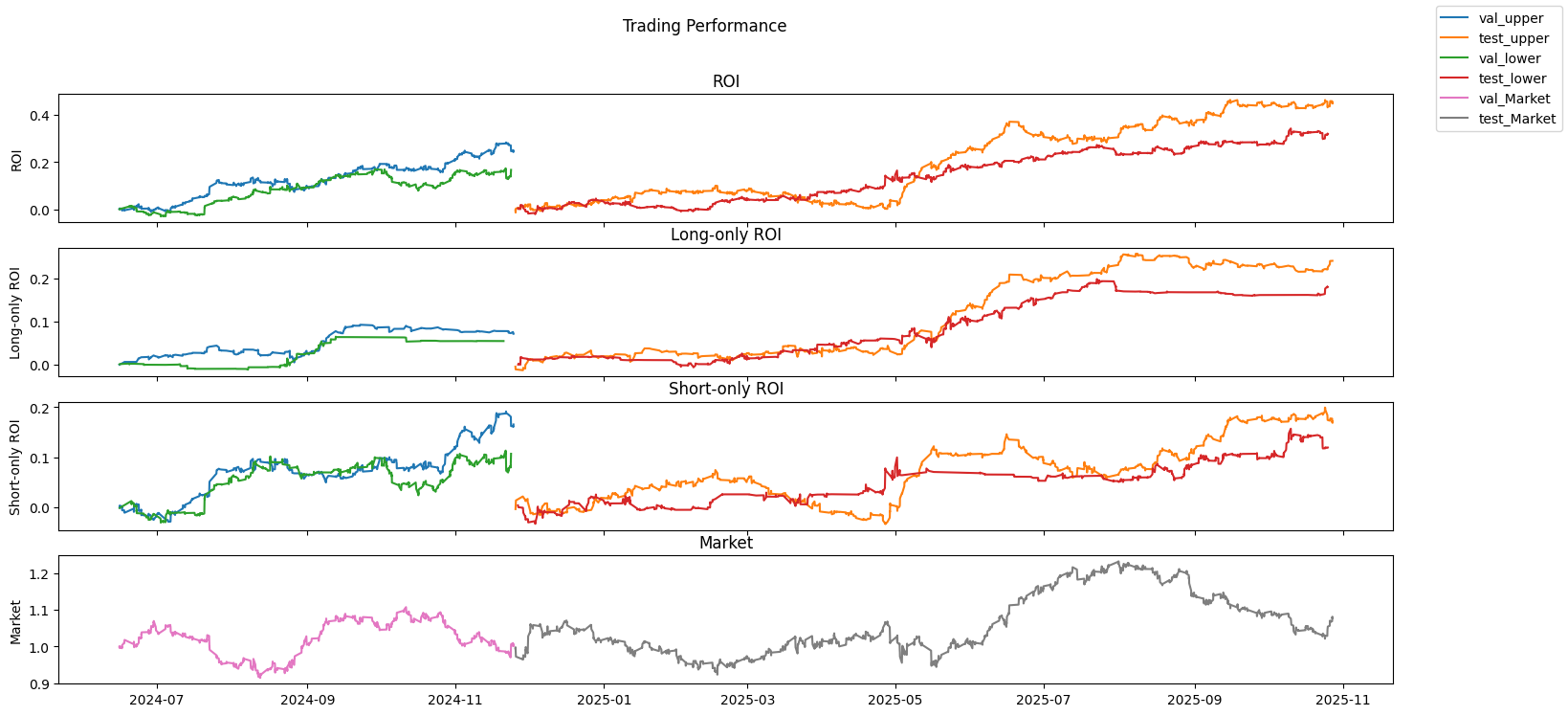}
    \caption{Strategy return curve of renewable energy}
    \label{fig:Renewable_Energy_ROI}
\end{figure}

\begin{figure}
    \centering
    \includegraphics[width=\linewidth]{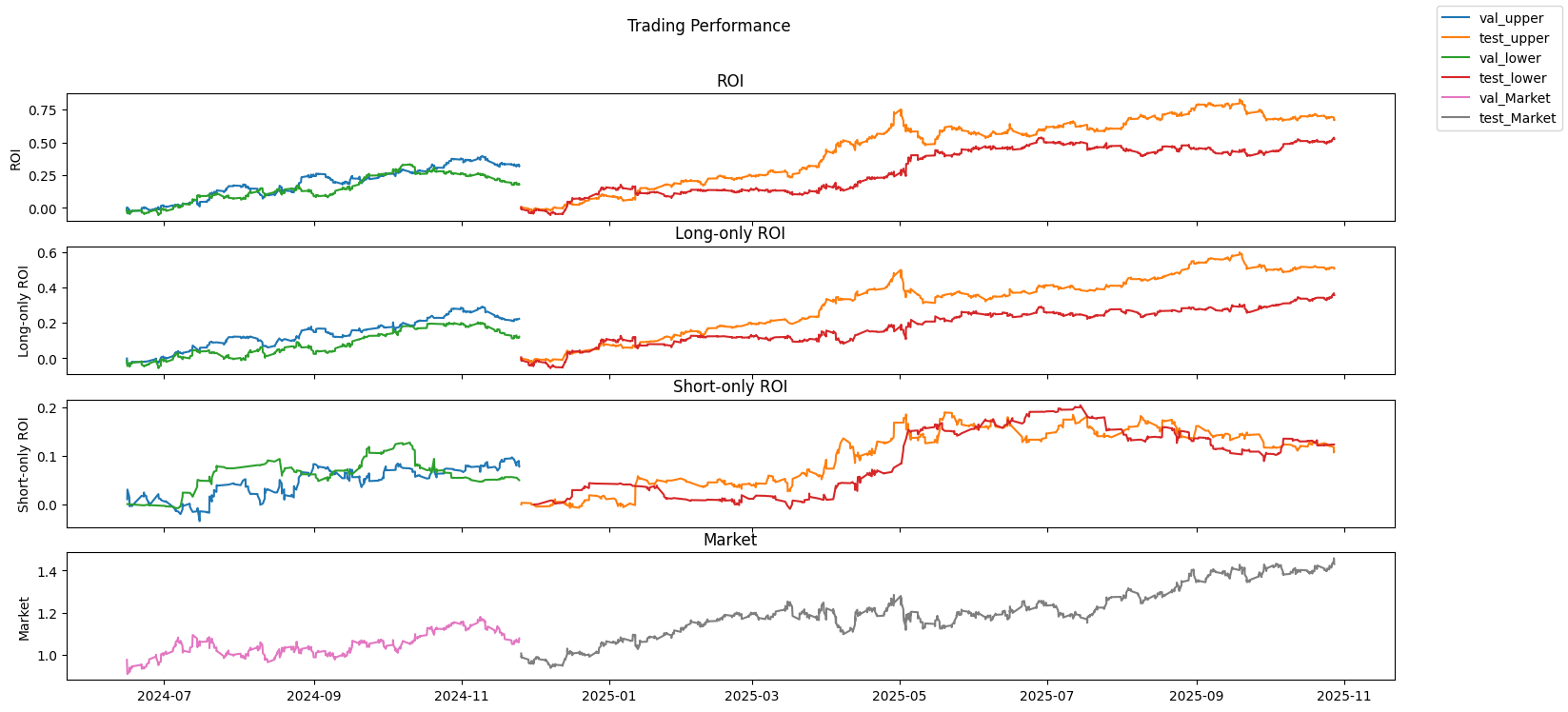}
    \caption{Strategy return curve of retail consumer goods}
    \label{fig:Retail_ConsumerGoods_ROI}
\end{figure}

\subsection{Hyperparameter Sensitivity Analysis}
\label{subsec:hyperparameter-sensitivity-analysis}
\subsubsection{$\lambda_\mathrm{roi}$ in Overfitting Penalty}
\label{subsec:lambda-in-roi-penalty}
The hyperparameter $\lambda_{\text{ROI}}$ controls the strength of the ROI-aware penalty and thus how aggressively the model is encouraged to maximize profit during training: small values make the constraint nearly inactive, whereas large values strongly discourage ROI-constraint violations.

\begin{figure}
\centering
\includegraphics[width=0.6\linewidth]{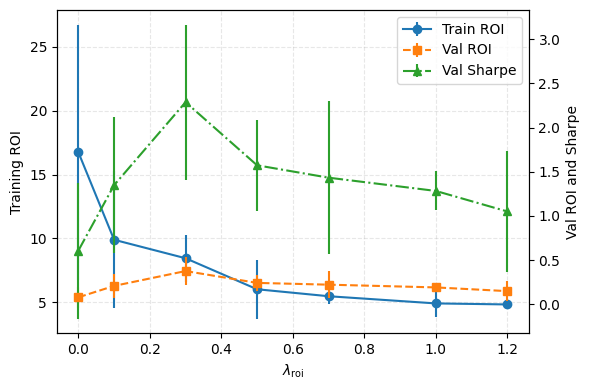}
\caption{Sensitivity of cumulative ROI to the ROI-aware penalty coefficient $\lambda_{\mathrm{roi}}$ on the Renewable Energy industry.
The left $y$-axis shows training cumulative ROI over the training horizon, and the right $y$-axis shows validation ROI and Sharpe.
All values are reported as mean$\pm$std over ten random seeds.}
\label{fig:lambda_roi}
\end{figure}

To analyze its effect, we sweep $\lambda_{\mathrm{roi}} \in \{0.0, 0.1, 0.3, 0.5, 0.7, 1.0, 1.2\}$, and train the model under the same configuration for each setting, and record the final training ROI over the full 4-year horizon together with the corresponding val ROI and Sharpe ratio. Results are shown in Fig.~\ref{fig:lambda_roi}.

We observe a clear dependence on $\lambda_{\mathrm{roi}}$. Too small a value yields weak regularization and overly aggressive policies that can produce inflated training ROI, while too large a value makes training overly conservative and substantially reduces ROI. In a mid-range of $\lambda_{\mathrm{roi}}$, training ROI stabilizes around 6 over the 4-year horizon and val ROI/Sharpe achieve their best values, which we adopt as a stable operating regime.

\subsubsection{$\gamma$ in Low-guided High-frequency Injection}
\label{subsec:gamma-in-lghi}
This subsection analyzes the scalar gate that controls the strength of LGHI. Empirically, overly large gate values cause vanishing gradients for LSTM backbones and exploding gradients for Transformer backbones, motivating a small-gate initialization.

Let $L_t \in \mathbb{R}^d$ be the low-frequency representation at time $t$ and $H_t$ the corresponding high-frequency input. LGHI produces a refinement $Z_t(L_t, H_t) \in \mathbb{R}^d$, and the fused output follows Eq.~\eqref{eq:gated_lowguided_injection}. Here, $\gamma$ is a learnable scalar and $\beta=\sigma(\gamma)\in(0,1)$ controls the overall contribution of LGHI: $\beta\approx 0$ yields an almost purely low-frequency representation, while $\beta\approx 1$ heavily relies on the high-frequency refinement.

The gate scales not only the forward contribution but also the backward signal to LGHI parameters $\theta_Z$. By the chain rule,
\begin{equation*}
  \frac{\partial \mathcal{L}}{\partial \theta_Z}
  =
  \frac{\partial \mathcal{L}}{\partial Y_t}\,
  \frac{\partial Y_t}{\partial Z_t}\,
  \frac{\partial Z_t}{\partial \theta_Z}
  =
  \beta\,
  \frac{\partial \mathcal{L}}{\partial Y_t}\,
  \frac{\partial Z_t}{\partial \theta_Z},
\end{equation*}
where $\mathcal{L}$ denotes the training objective. Thus, larger $\beta$ linearly amplifies gradients w.r.t.\ $\theta_Z$, while also increasing the scale of $Y_t$ fed into the backbone.

\noindent\textbf{Effect with LSTM backbones}
With an LSTM backbone, $Y_t$ enters recurrent transitions whose gates use sigmoid and $\tanh$. As $\beta$ increases, the magnitude of $Y_t$ tends to grow, pushing gate pre-activations into saturation where $\sigma'(x)$ and $1-\tanh^2(x)$ become small. Consequently, the recurrent-step Jacobian tends to have a spectral radius below $1$, and the backpropagated gradients decay through time:
\begin{equation*}
  \frac{\partial \mathcal{L}}{\partial (h_t, c_t)}
  =
  \Bigg(
    \prod_{k=t}^{T-1} J_k
  \Bigg)
  \frac{\partial \mathcal{L}}{\partial (h_T, c_T)},
\end{equation*}
leading to vanishing gradients and early training plateaus for moderately large $\beta$.

\begin{figure}
	\centering
	\includegraphics[width=\linewidth]{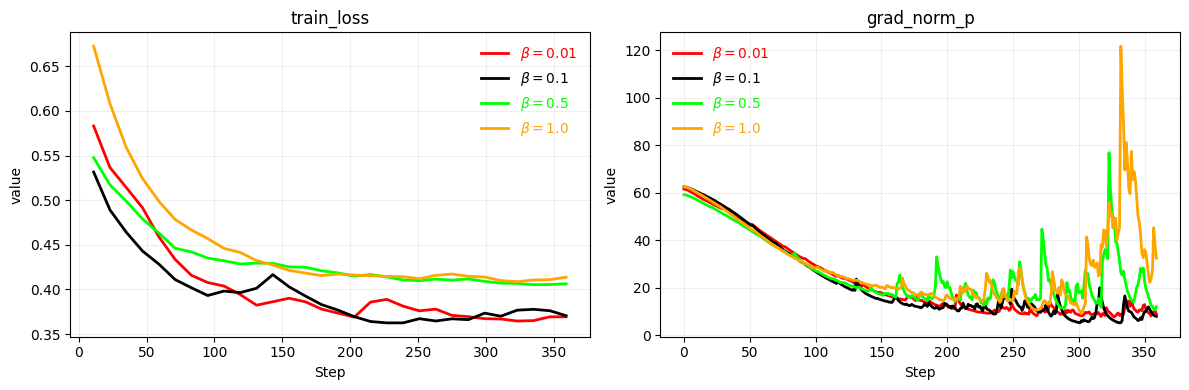}
	\caption{Effect of the LGHI module $\beta$ on training loss and gradient norms for Transformer backbones.}
	\label{fig:beta-gradient}
\end{figure}

\noindent\textbf{Effect with Transformer backbones}
In contrast, a Transformer backbone forms a deep stack of residual blocks with Layer Normalization $X^{(\ell+1)} = \mathrm{LN}\bigl(X^{(\ell)} + F^{(\ell)}(X^{(\ell)})\bigr)$.
A first-order approximation yields a layer Jacobian of the form $J^{(\ell)} \approx I + \beta\, \tilde{J}^{(\ell)}$.
Thus, increasing $\beta$ raises the effective layer-wise gain, and the backward signal can grow through the product of Jacobians:
\begin{equation*}
  \frac{\partial \mathcal{L}}{\partial X^{(0)}}
  =
  \Bigg(
    \prod_{\ell=0}^{L-1} J^{(\ell)}
  \Bigg)
  \frac{\partial \mathcal{L}}{\partial X^{(L)}},
\end{equation*}
which may cause exploding gradients when the spectral radius exceeds $1$ across many layers. In our experiments, setting $\beta \in \{0.01, 0.1, 0.5, 1.0\}$ shows stable training for small/moderate gates, while $\beta=0.5$ and $\beta=1.0$ quickly become numerically unstable as shown in Fig.~\ref{fig:beta-gradient}.

\noindent\textbf{Initialization strategy}
Across backbones, overly large gates can stall optimization by saturating LSTM gates and causing vanishing gradients, or by over-amplifying deep residual stacks in Transformers and triggering exploding gradients. Therefore, we initialize $\beta$ to a very small value so that LGHI starts as a weak perturbation of the low-frequency path. Specifically, we parameterize $\beta = \sigma(\gamma)$ and initialize $\gamma = -5$, yielding $\beta_0 = \sigma(-5) \approx 0.0067$. This keeps the mapping close to the identity at initialization and avoids both LSTM saturation and excessive amplification in deep Transformer stacks. Since $\gamma$ is learnable, the optimizer can increase $\beta$ when the refinement $Z_t(L_t,H_t)$ is beneficial; otherwise, $\beta$ remains small and the model behaves close to a purely low-frequency baseline.

\subsubsection{$\lambda_{\mathrm{spec}}$ in Neural Wavelet Loss}
The hyperparameter $\lambda_{\mathrm{spec}}$ weights the spectral-shaping term in the wavelet loss.
Smaller values relax frequency-domain constraints and allow more task-driven filters, while larger values enforce cleaner low-/high-pass behavior but can reduce flexibility.

\begin{figure}
\centering
\includegraphics[width=0.6\linewidth]{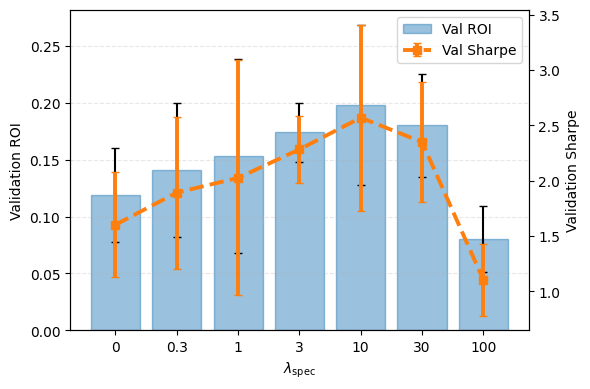}
\caption{Validation-set sensitivity of $\lambda_{\mathrm{spec}}$ in the wavelet loss on the Renewable Energy industry. Bars report validation ROI on the left $y$-axis, and the dashed line reports validation Sharpe ratio on the right $y$-axis. All values are reported as mean$\pm$std over ten random seeds.}
\label{fig:lambda_spec}
\end{figure}

We sweep $\lambda_{\mathrm{spec}} \in \{0,\ 0.3,\ 1,\ 3,\ 10,\ 30,\ 100\}$, keeping all other settings fixed, and evaluate validation ROI and Sharpe.
Fig.~\ref{fig:lambda_spec} summarizes the results on Renewable Energy.
We select $\lambda_{\mathrm{spec}}$ using validation performance to avoid test-set leakage, and report test results with the selected value in the main results section.

Overall, performance peaks in a moderate range.
Too small $\lambda_{\mathrm{spec}}$ weakens spectral shaping and can yield less stable filters, while too large $\lambda_{\mathrm{spec}}$ over-regularizes optimization and degrades ROI and Sharpe.
We use $\lambda_{\mathrm{spec}}=10$ as the default unless otherwise stated.

\subsection{Ablation Study}
\label{subsec:ablation-study}
\subsubsection{Effect of Learnable Wavelet Filters}
We evaluate the proposed neural wavelet layer by replacing fixed filters in the same Transformer backbone. We compare three front-ends: (i) \emph{Neural Wavelet}, which learns the filters jointly with the trading objective using Parseval, overlap, and pass-band regularizers; (ii) \emph{Classic Wavelet}, a fixed wavelet transform; and (iii) \emph{Transformer only}, which uses raw price features. As shown in Table~\ref{tab:wavelet_transformer_multiseed}, Neural Wavelet attains the best test ROI and Sharpe in almost all industries under matched capacity, and also yields smaller maximum draw-down and more stable performance across seeds, indicating that learnable, regularized wavelet filters extract more tradable signals than fixed transforms or no decomposition.

\begin{table}[!h]
  \centering
  \footnotesize
  \setlength{\tabcolsep}{4pt}
  \caption{Multi-seed, multi-industry comparison of Transformer-based backbones with different front-ends. Results are reported as mean $\pm$ std over ten seeds. Best results in each row are highlighted in \textbf{bold}.}
  \label{tab:wavelet_transformer_multiseed}
  \renewcommand{\arraystretch}{1}
  \begin{tabular}{llccc}
    \toprule
    Industry & Metric & \makecell{Neural wavelet\\Transformer} & \makecell{Classic wavelet\\Transformer} & \makecell{Plain\\Transformer} \\
    \midrule 
    \multirow{2}{*}{Biotechnology} 
    & ROI & $\bm{0.601 \pm 0.034}$ & 0.323 $\pm$ 0.021 & 0.124 $\pm$ 0.066 \\
    & Sharpe & $\bm{1.695 \pm 0.050}$ & 1.086 $\pm$ 0.048 & 0.511 $\pm$ 0.189 \\
    \midrule
    \multirow{2}{*}{Semiconductors} 
    & ROI & $\bm{1.104 \pm 0.053}$ & 0.416 $\pm$ 0.021 & 0.333 $\pm$ 0.021 \\
    & Sharpe & $\bm{2.555 \pm 0.081}$ & 1.293 $\pm$ 0.071 & 1.134 $\pm$ 0.035 \\
    \midrule
    \multirow{2}{*}{Renewable Energy} 
    & ROI & $\bm{0.423 \pm 0.074}$ & 0.261 $\pm$ 0.033 & 0.169 $\pm$ 0.018 \\
    & Sharpe & $\bm{2.775 \pm 0.365}$ & 1.723 $\pm$ 0.161 & 1.173 $\pm$ 0.113 \\
    \midrule
    \multirow{2}{*}{Life Insurance}
    & ROI & $\bm{0.185 \pm 0.004}$ & 0.144 $\pm$ 0.004 & 0.120 $\pm$ 0.010 \\
    & Sharpe & $\bm{1.472 \pm 0.039}$ & 1.206 $\pm$ 0.027 & 0.976 $\pm$ 0.086 \\
    \midrule
    \multirow{2}{*}{Medical Devices}
    & ROI & $\bm{0.669 \pm 0.049}$ & 0.462 $\pm$ 0.067 & 0.289 $\pm$ 0.036 \\
    & Sharpe & $\bm{1.979 \pm 0.167}$ & 1.475 $\pm$ 0.178 & 1.060 $\pm$ 0.125 \\
    \midrule
    \multirow{2}{*}{Retail Consumer Goods}   
    & ROI    & $\bm{0.659 \pm 0.055}$ & 0.468 $\pm$ 0.143 & 0.317 $\pm$ 0.187 \\
    & Sharpe & $\bm{2.465 \pm 0.293}$ & 1.850 $\pm$ 0.132 & 1.291 $\pm$ 0.185 \\
    \midrule
    \multirow{2}{*}{Overall (avg.)} 
    & ROI    & $\bm{0.607 \pm 0.045}$ & 0.346 $\pm$ 0.048 & 0.225 $\pm$ 0.056 \\
    & Sharpe & $\bm{2.157 \pm 0.166}$ & 1.439 $\pm$ 0.103 & 1.024 $\pm$ 0.122 \\
    \bottomrule
  \end{tabular}
\end{table}

\subsubsection{Effect of Using Only a Single Frequency Branch}
We evaluate whether both frequency branches are necessary by comparing the full model (Low+High) with two ablations: \emph{Low-Freq only} and \emph{High-Freq only}. Table~\ref{tab:freq_branch_multiseed} shows that removing either branch consistently hurts performance. Across most industries, \emph{Low-Freq only} outperforms \emph{High-Freq only} in ROI and Sharpe, indicating that slowly varying trends are more informative than high-frequency residuals alone. However, Low+High achieves the best average ROI/Sharpe with lower seed variance, suggesting that the high-frequency branch provides complementary short-term refinements, whereas relying solely on high-frequency signals reduces mean performance and increases training instability.

\begin{table}[!h]\centering\footnotesize\setlength{\tabcolsep}{4pt}
\caption{Effect of two frequency branches. Results are mean $\pm$ std over ten seeds. Best in each row is \textbf{bold}.}
\label{tab:freq_branch_multiseed}
\renewcommand{\arraystretch}{1}
\begin{tabular}{llccc}
\toprule
Industry & Metric & \makecell{Full\\Low+High} & \makecell{Only\\Low-Freq} & \makecell{Only\\High-Freq}\\
\midrule
\multirow{2}{*}{Biotechnology}  
& ROI    & $\bm{0.601 \pm 0.034}$ & 0.305 $\pm$ 0.028 & 0.133 $\pm$ 0.050\\
& Sharpe & $\bm{1.695 \pm 0.050}$ & 0.902 $\pm$ 0.117 & 0.429 $\pm$ 0.185\\
\midrule 
\multirow{2}{*}{Semiconductors} 
& ROI    & $\bm{1.104 \pm 0.053}$ & 0.561 $\pm$ 0.052 & 0.244 $\pm$ 0.092\\
& Sharpe & $\bm{2.555 \pm 0.081}$ & 1.360 $\pm$ 0.176 & 0.646 $\pm$ 0.279\\
\midrule 
\multirow{2}{*}{Renewable Energy}
& ROI    & $\bm{0.423 \pm 0.074}$ & 0.225 $\pm$ 0.021 & 0.098 $\pm$ 0.037\\
& Sharpe & $\bm{2.775 \pm 0.365}$ & 1.477 $\pm$ 0.191 & 0.702 $\pm$ 0.303\\
\midrule 
\multirow{2}{*}{Life Insurance}    
& ROI    & $\bm{0.185 \pm 0.004}$ & 0.094 $\pm$ 0.009 & 0.041 $\pm$ 0.015\\
& Sharpe & $\bm{1.472 \pm 0.039}$ & 0.783 $\pm$ 0.101 & 0.372 $\pm$ 0.161\\
\midrule 
\multirow{2}{*}{Medical Devices} 
& ROI    & $\bm{0.669 \pm 0.049}$ & 0.340 $\pm$ 0.032 & 0.148 $\pm$ 0.056\\
& Sharpe & $\bm{1.979 \pm 0.167}$ & 1.053 $\pm$ 0.136 & 0.501 $\pm$ 0.216\\
\midrule 
\multirow{2}{*}{Retail Consumer Goods}     
& ROI    & $\bm{0.659 \pm 0.055}$ & 0.334 $\pm$ 0.165 & 0.145 $\pm$ 0.286\\
& Sharpe & $\bm{2.465 \pm 0.293}$ & 1.312 $\pm$ 0.161 & 0.624 $\pm$ 0.329\\
\midrule 
\multirow{2}{*}{Overall (avg.)}    
& ROI     & $\bm{0.607 \pm 0.045}$ & 0.310 $\pm$ 0.051 & 0.135 $\pm$ 0.089\\
& Sharpe  & $\bm{2.157 \pm 0.166}$ & 1.148 $\pm$ 0.147 & 0.546 $\pm$ 0.246\\

\bottomrule
\end{tabular}
\end{table}

\subsubsection{Low-guided High-frequency Injection}
We further examine how the low and high frequency branches should be fused. In the proposed design, the low-frequency representation attends to the high-frequency branch via a LGHI block. As a comparison, we remove the LGHI module and simply concatenate the two branches followed by a linear projection, while keeping the overall model size comparable. The multi-industry, multi-seed results are reported in Table~\ref{tab:lghi_vs_concat}.

\begin{table}[!h]\centering\footnotesize\setlength{\tabcolsep}{8pt}
\caption{LGHI module vs simple concatenation. Results are mean $\pm$ std over ten seeds. Best in each row is \textbf{bold}.}
\label{tab:lghi_vs_concat}
\renewcommand{\arraystretch}{1}
\begin{tabular}{llcc}
\toprule
Industry & Metric & LGHI & Concat\\
\midrule
\multirow{2}{*}{Biotechnology}  
& ROI    & $\bm{0.601 \pm 0.034}$ & 0.203 $\pm$ 0.037\\
& Sharpe & $\bm{1.695 \pm 0.050}$ & 0.640 $\pm$ 0.040\\
\midrule 
\multirow{2}{*}{Semiconductors} 
& ROI    & $\bm{1.104 \pm 0.053}$ & 0.374 $\pm$ 0.022\\
& Sharpe & $\bm{2.555 \pm 0.081}$ & 0.964 $\pm$ 0.060\\
\midrule 
\multirow{2}{*}{Renewable Energy} 
& ROI    & $\bm{0.423 \pm 0.074}$ & 0.150 $\pm$ 0.035\\
& Sharpe & $\bm{2.775 \pm 0.365}$ & 1.047 $\pm$ 0.165\\
\midrule 
\multirow{2}{*}{Life Insurance}     
& ROI    & $\bm{0.185 \pm 0.004}$ & 0.063 $\pm$ 0.012\\
& Sharpe & $\bm{1.472 \pm 0.039}$ & 0.555 $\pm$ 0.034\\
\midrule 
\multirow{2}{*}{Medical Devices}     
& ROI    & $\bm{0.669 \pm 0.049}$ & 0.227 $\pm$ 0.028\\
& Sharpe & $\bm{1.979 \pm 0.167}$ & 0.747 $\pm$ 0.076\\
\midrule 
\multirow{2}{*}{Retail Consumer Goods}
& ROI & $\bm{0.659 \pm 0.055}$ & 0.222 $\pm$ 0.154\\
& Sharpe & $\bm{2.465 \pm 0.293}$ & 0.930 $\pm$ 0.130\\
\midrule 
\multirow{2}{*}{Overall (avg.)}
& ROI    & $\bm{0.607 \pm 0.045}$ & 0.207 $\pm$ 0.048\\
& Sharpe & $\bm{2.157 \pm 0.166}$ & 0.814 $\pm$ 0.083\\
\bottomrule
\end{tabular}
\end{table}

Across all industry groups, the LGHI design consistently outperforms simple concatenation in terms of both ROI and Sharpe ratio, and also exhibits smaller variance across seeds on average. This confirms that allowing the low-frequency branch to selectively attend to high-frequency details is more effective than treating the two branches as independent features.

\subsubsection{Sharpe Regularizer Ablation}
\label{sec:sharpe-loss-ablation}
To quantify the benefit of explicitly optimizing a trading-oriented objective, we ablate the Sharpe loss term by training two WaveLSFormer variants with identical architectures, optimization settings, and data splits: (i) Soft-label only and (ii) Soft-label with a Sharpe loss computed from per-step P\&L. 
For each industry and seed, both variants share the same training schedule and differ only in whether the Sharpe term is included. 
Table~\ref{tab:sharpe_loss_ablation_industry} reports industry-level test metrics averaged over 10 seeds with checkpoints selected by validation ROI. Adding the Sharpe loss improves risk-adjusted performance and often increases ROI while reducing maximum draw-down, suggesting that prediction-only training is insufficient and that directly regularizing the mean-volatility trade-off yields more favorable risk-return profiles.

Beyond ROI and Sharpe ratio, we report maximum draw-down (MDD) on the test set as a complementary risk measure. 
Let $W_t$ denote the cumulative wealth process. The draw-down at time $t$ and the maximum draw-down over the test period are
\begin{equation*}
  D_t = 1 - \frac{W_t}{\max_{s \le t} W_s}, 
  \quad
  \mathrm{MDD} = \max_t D_t.
\end{equation*}

\begin{figure}
    \centering
    \includegraphics[width=\linewidth]{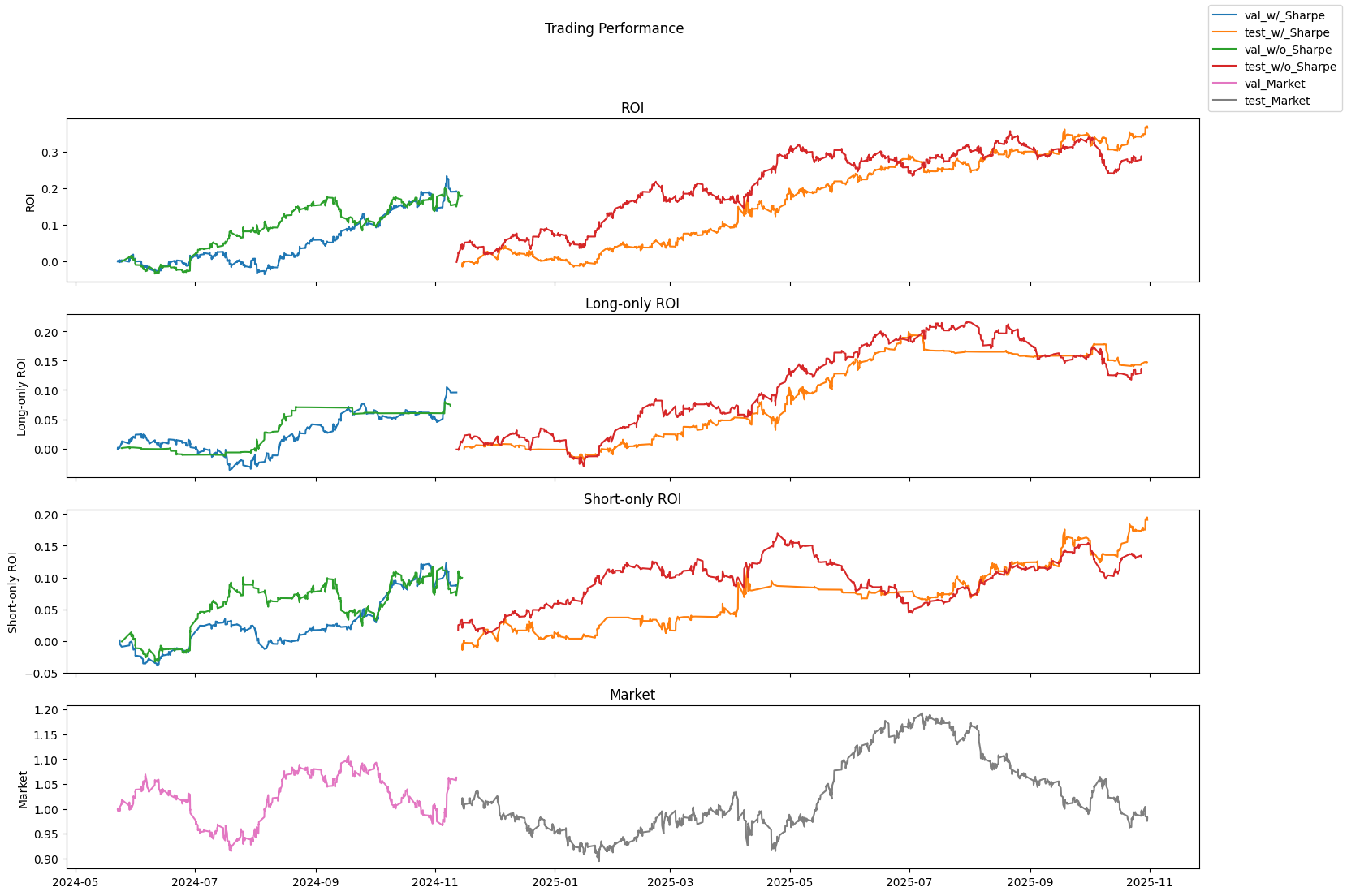}
    \caption{Equity curves with and without the Sharpe loss for Renewable Energy. Top: cumulative ROI on validation and test. Bottom: cumulative market return. The model without Sharpe loss shows deeper and more frequent test draw-downs, whereas the Sharpe-aware model produces a smoother equity curve with smaller draw-downs at a similar final ROI, reflecting more conservative exposure in risky periods.}
    \label{fig:Renewable_Energy_wo_sharpe}
\end{figure}

\begin{table}[!h]\centering\setlength{\tabcolsep}{3pt}\footnotesize
  \centering
  \caption{Sharpe loss ablation by industry on the test set. Results are reported as mean $\pm$ standard deviation over 10 random seeds for ROI, Sharpe ratio and maximum draw-down (MDD, \%). The best overall results are in \textbf{bold}}
  \label{tab:sharpe_loss_ablation_industry}
  \renewcommand{\arraystretch}{1}
  \resizebox{\linewidth}{!}{%
  \begin{tabular}{lcccccc}
    \toprule
    \multirow{2}{*}{Industry}
      & \multicolumn{3}{c}{Soft-label only}
      & \multicolumn{3}{c}{Soft-label + Sharpe (ours)} \\
    \cmidrule(lr){2-4} \cmidrule(lr){5-7}
      & ROI & Sharpe & MDD (\%) & ROI & Sharpe & MDD (\%) \\
    \midrule
    Biotechnology         
      & 0.511 $\pm$ 0.061 & 1.187 $\pm$ 0.190 & 17.203 $\pm$ 2.962
      & 0.601 $\pm$ 0.034 & 1.695 $\pm$ 0.050 & 10.658 $\pm$ 1.932 \\
    Semiconductors        
      & 0.938 $\pm$ 0.113 & 1.789 $\pm$ 0.286 & 11.413 $\pm$ 1.965
      & 1.104 $\pm$ 0.053 & 2.555 $\pm$ 0.081 &  7.071 $\pm$ 1.282 \\
    Renewable Energy      
      & 0.377 $\pm$ 0.045 & 1.943 $\pm$ 0.311 & 10.508 $\pm$ 1.809
      & 0.443 $\pm$ 0.064 & 2.775 $\pm$ 0.365 &  6.510 $\pm$ 1.180 \\
    Life Insurance        
      & 0.157 $\pm$ 0.019 & 1.030 $\pm$ 0.165 & 19.810 $\pm$ 3.410
      & 0.185 $\pm$ 0.004 & 1.472 $\pm$ 0.039 & 12.273 $\pm$ 2.225 \\
    Medical Devices       
      & 0.569 $\pm$ 0.068 & 1.385 $\pm$ 0.222 & 14.735 $\pm$ 2.537
      & 0.669 $\pm$ 0.049 & 1.979 $\pm$ 0.167 &  9.128 $\pm$ 1.655 \\
    Retail Consumer Goods 
      & 0.267 $\pm$ 0.032 & 1.068 $\pm$ 0.171 & 19.109 $\pm$ 3.290
      & 0.659 $\pm$ 0.055 & 2.465 $\pm$ 0.293 & 11.838 $\pm$ 2.146 \\

    \midrule
    Overall (avg.)
      & 0.470 $\pm$ 0.056 & 1.400 $\pm$ 0.224 & 15.463 $\pm$ 2.662
      & $\bm{0.553 \pm 0.035}$ & $\bm{2.000 \pm 0.122}$ & $\bm{9.580 \pm 1.736}$ \\
    \bottomrule
  \end{tabular}%
  }
\end{table}

\subsection{Baseline Model Comparisons}
\label{subsec:baseline-model-comparisons}
\subsubsection{Per-Industry Performance}
We compare MLP, LSTM, and Transformer backbones with and without the neural wavelet front-end, using $d_{\text{model}}=512$ for all models.
WaveLSFormer uses $d_{\text{ff}}=1024$, $n_{\text{heads}}=128$, $L=6$ encoder layers, sequence length $96$, temporal embedding dimension $128$, and pre-layer normalization.
The MLP baseline has 10 fully connected layers with 512 hidden units, and the LSTM baseline uses a 2-layer LSTM with 512 hidden units.
We report mean$\pm$std over ten random seeds and synchronize stochastic components across models for fair comparison.

\begin{figure}
    \centering
    \includegraphics[width=0.95\linewidth]{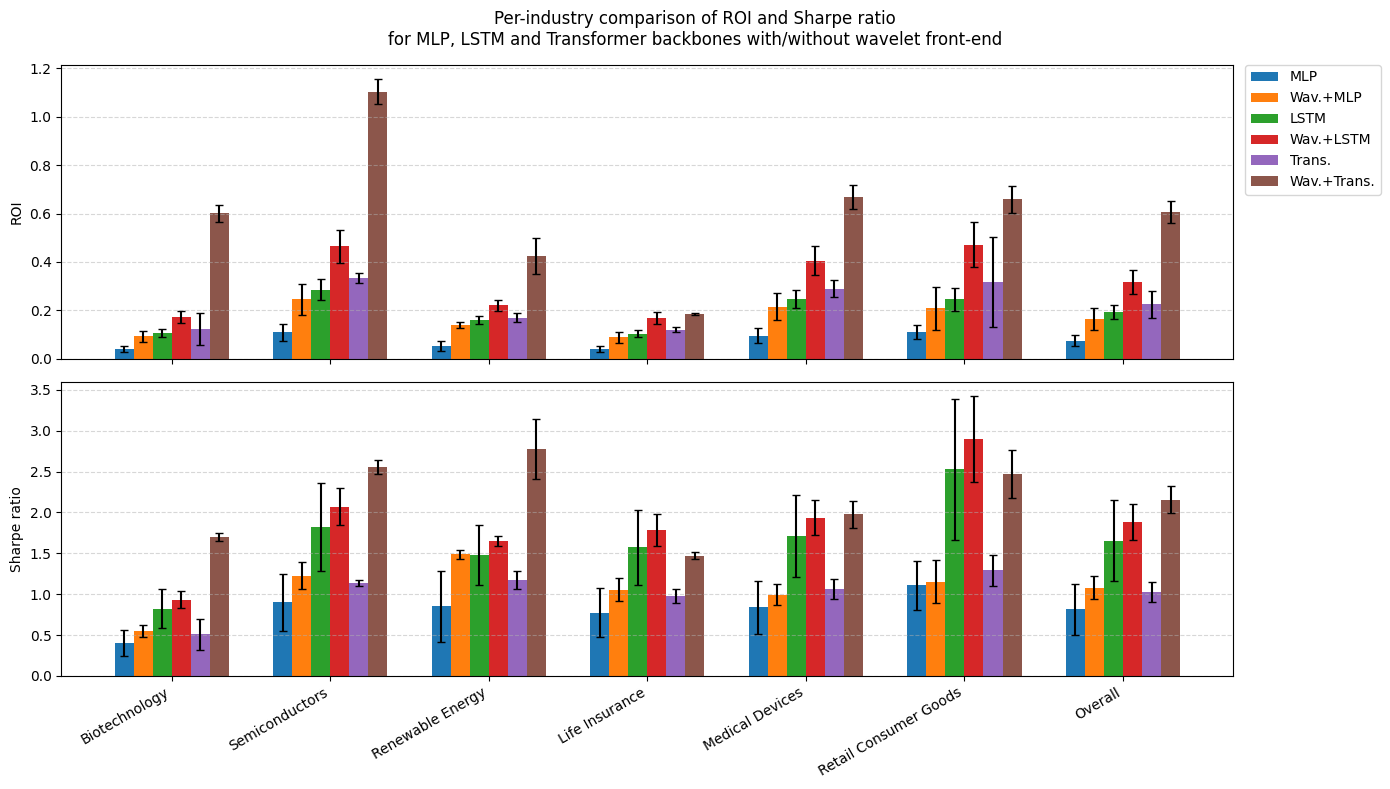}
    \caption{Per-industry ROI (a) and Sharpe ratio (b) for MLP, LSTM, and Transformer backbones with and without the neural wavelet front-end. Bars show mean over ten seeds and error bars denote one standard deviation, including six industries and the overall average.}
    \label{fig:backbone_bar}
\end{figure}

\begin{figure}
    \centering
    \begin{subfigure}{\linewidth}
        \centering
        \includegraphics[width=\linewidth]{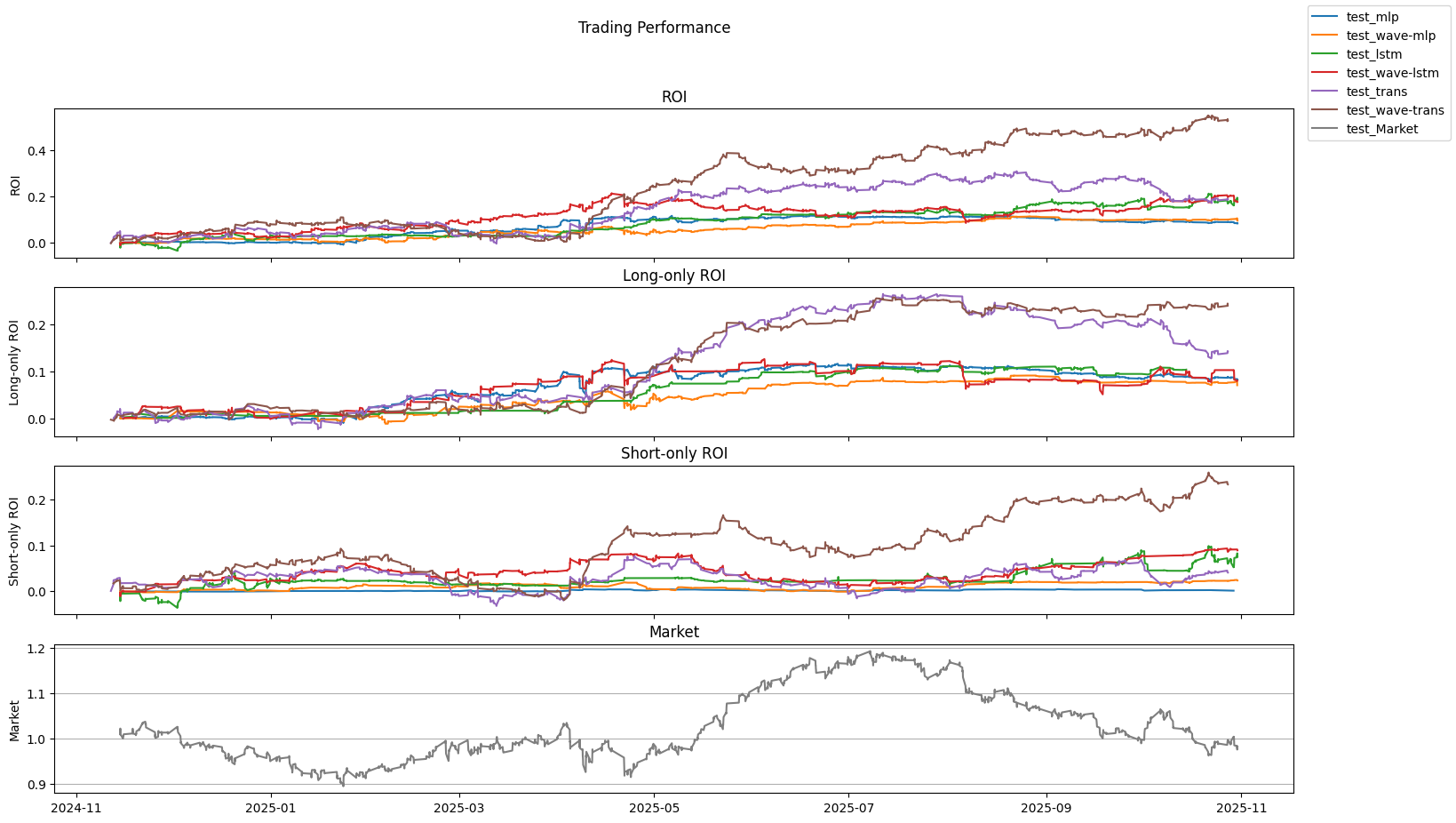}
        \caption{Renewable Energy}
    \end{subfigure}
    \vspace{0.3em}
    \begin{subfigure}{\linewidth}
        \centering
        \includegraphics[width=\linewidth]{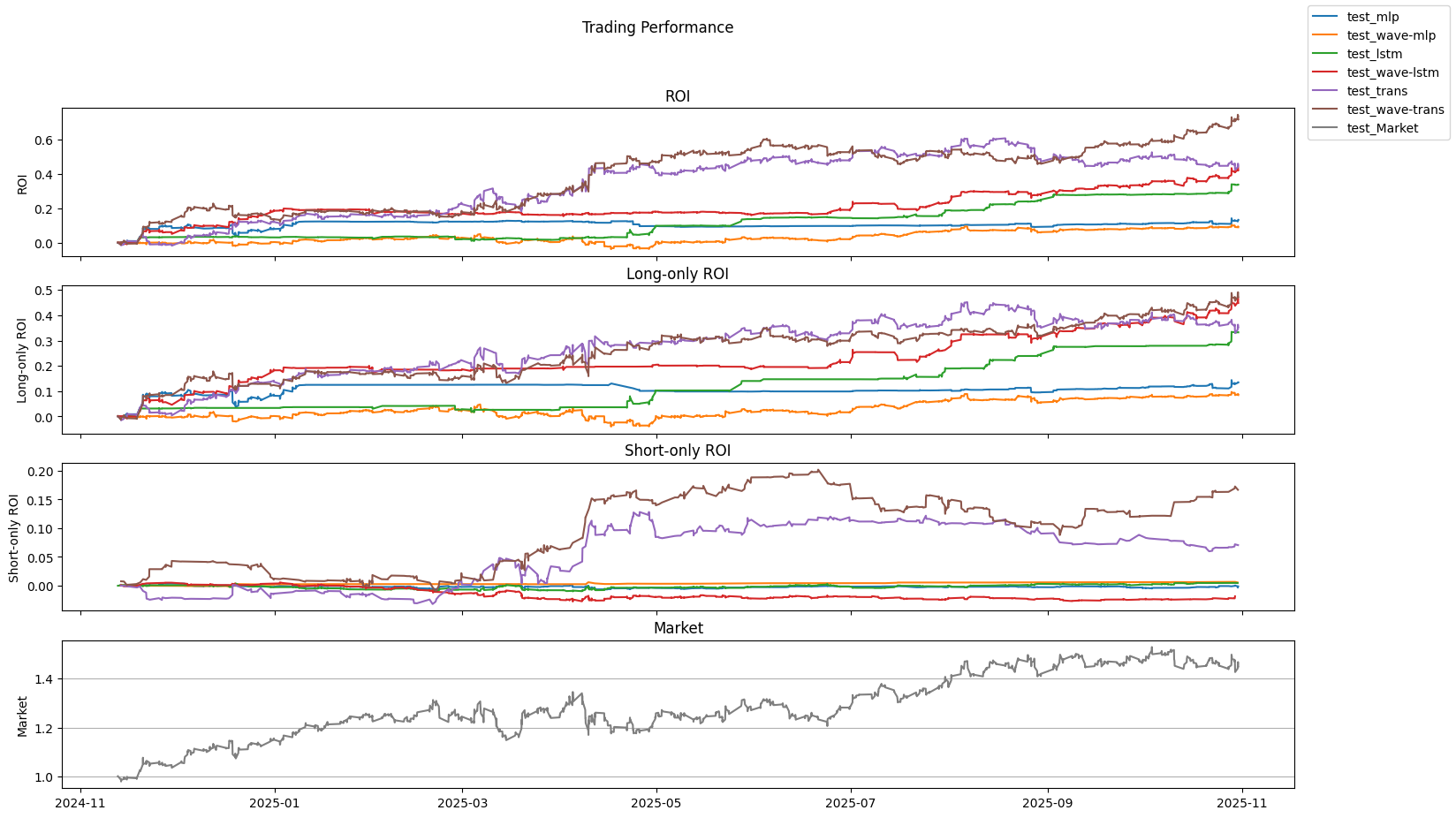}
        \caption{Retail Consumer Goods}
    \end{subfigure}
    \caption{Equity curves on the test period for (a) Renewable Energy and (b) Retail Consumer Goods. For each sector we compare the non-wavelet MLP and LSTM, the wavelet-enhanced MLP and LSTM, and WaveLSFormer.}
    \label{fig:equity_re_rcg}
\end{figure}

Table~\ref{tab:overall_backbones} reports test ROI and Sharpe by industry and on average.
The wavelet front-end improves both metrics for every backbone, and LSTM-based models consistently outperform MLP, consistent with prior evidence that recurrent and attention architectures better capture long-range dependencies and regime shifts in financial time series~\cite{Nelson2017LSTM, Moghar2020LSTM, Bao2017DLF, Kabir2025LSTMTrans, Mozaffari2024Transformer}.
Overall, WaveLSFormer achieves the best performance with a parameter budget comparable to LSTM variants, reaching $0.607 \pm 0.045$ ROI and $2.157 \pm 0.166$ Sharpe, compared with $0.225 \pm 0.056$ ROI and $1.024 \pm 0.122$ Sharpe for the plain Transformer.
Fig.~\ref{fig:backbone_bar} shows consistent gains across industries, and Fig.~\ref{fig:equity_re_rcg} shows smoother equity growth with smaller draw-downs than the strongest non-wavelet baselines in Renewable Energy and Retail Consumer Goods.

\begin{table}[h!]\centering\setlength{\tabcolsep}{2pt}\footnotesize
\caption{Per-industry comparison of MLP, LSTM and Transformer backbones with and without the wavelet front-end. Results are mean $\pm$ std over ten seeds. The best results in each row are in \textbf{bold}.}
\label{tab:overall_backbones}
\renewcommand{\arraystretch}{1}
\resizebox{\linewidth}{!}{%
\begin{tabular}{llcccccc}
\toprule
Industry & Metric & MLP & Wav.+MLP & LSTM & Wav.+LSTM & Trans. & Wav.+Trans.\\
\midrule
\multirow{2}{*}{Biotechnology} 
  & ROI    & $0.041 \pm 0.013$ & $0.092 \pm 0.024$ & $0.106 \pm 0.016$ & $0.173 \pm 0.026$ & $0.124 \pm 0.066$ & $\bm{0.601 \pm 0.034}$\\
  & Sharpe & $0.405 \pm 0.157$ & $0.553 \pm 0.074$ & $0.823 \pm 0.242$ & $0.934 \pm 0.102$ & $0.511 \pm 0.189$ & $\bm{1.695 \pm 0.050}$\\

\multirow{2}{*}{Semiconductors} 
  & ROI    & $0.110 \pm 0.035$ & $0.246 \pm 0.064$ & $0.285 \pm 0.043$ & $0.465 \pm 0.069$ & $0.333 \pm 0.021$ & $\bm{1.104 \pm 0.053}$\\
  & Sharpe & $0.899 \pm 0.349$ & $1.227 \pm 0.164$ & $1.826 \pm 0.537$ & $2.072 \pm 0.227$ & $1.134 \pm 0.035$ & $\bm{2.555 \pm 0.081}$\\

\multirow{2}{*}{Renewable Energy} 
  & ROI    & $0.053 \pm 0.020$ & $0.139 \pm 0.013$ & $0.159 \pm 0.017$ & $0.221 \pm 0.022$ & $0.169 \pm 0.018$ & $\bm{0.423 \pm 0.074}$\\
  & Sharpe & $0.851 \pm 0.431$ & $1.490 \pm 0.054$ & $1.482 \pm 0.366$ & $1.656 \pm 0.061$ & $1.173 \pm 0.113$ & $\bm{2.775 \pm 0.365}$\\

\multirow{2}{*}{Life Insurance} 
  & ROI    & $0.040 \pm 0.013$ & $0.089 \pm 0.023$ & $0.103 \pm 0.015$ & $0.168 \pm 0.025$ & $0.120 \pm 0.010$ & $\bm{0.185 \pm 0.004}$\\
  & Sharpe & $0.774 \pm 0.301$ & $1.056 \pm 0.141$ & $1.572 \pm 0.462$ & $\bm{1.783 \pm 0.195}$ & $0.976 \pm 0.086$ & $1.472 \pm 0.039$\\

\multirow{2}{*}{Medical Devices} 
  & ROI    & $0.095 \pm 0.030$ & $0.214 \pm 0.056$ & $0.248 \pm 0.037$ & $0.404 \pm 0.060$ & $0.289 \pm 0.036$ & $\bm{0.669 \pm 0.049}$\\
  & Sharpe & $0.840 \pm 0.327$ & $0.995 \pm 0.133$ & $1.707 \pm 0.502$ & $1.937 \pm 0.212$ & $1.060 \pm 0.125$ & $\bm{1.979 \pm 0.167}$\\

\multirow{2}{*}{Retail Consumer Goods} 
  & ROI    & $0.110 \pm 0.028$ & $0.208 \pm 0.089$ & $0.245 \pm 0.047$ & $0.471 \pm 0.093$ & $0.317 \pm 0.187$ & $\bm{0.659 \pm 0.055}$\\
  & Sharpe & $1.110 \pm 0.301$ & $1.154 \pm 0.267$ & $2.527 \pm 0.862$ & $\bm{2.895 \pm 0.528}$ & $1.291 \pm 0.185$ & $2.465 \pm 0.293$\\

\midrule
\multirow{2}{*}{Overall (avg.)} 
  & ROI    & $0.075 \pm 0.023$ & $0.165 \pm 0.045$ & $0.191 \pm 0.029$ & $0.317 \pm 0.049$ & $0.225 \pm 0.056$ & $\bm{0.607 \pm 0.045}$\\
  & Sharpe & $0.813 \pm 0.311$ & $1.079 \pm 0.139$ & $1.656 \pm 0.495$ & $1.879 \pm 0.221$ & $1.024 \pm 0.122$ & $\bm{2.157 \pm 0.166}$\\
\bottomrule
\end{tabular}%
}
\end{table}

\subsubsection{Model Complexity and Overall Performance}
Table~\ref{tab:overall_baselines} reports model complexity and trading performance for each backbone with and without the neural wavelet front-end.
MLP is a lightweight reference with $8.146$M parameters and $3.468$G FLOPs, while LSTM and Transformer are substantially more expensive at $12.538$M/\allowbreak$492.614$G and $15.928$M/\allowbreak$665.921$G.
To rule out brute-force scaling, we report parameters and FLOPs for paired comparisons within each backbone.
The wavelet module introduces only small capacity changes, increasing parameters to $8.151$M for MLP, $13.298$M for LSTM, and $15.943$M for Transformer.
Its computational overhead is modest, with LSTM FLOPs increasing by $7.8\%$ and Transformer FLOPs slightly decreasing by $0.9\%$.

At comparable complexity, the wavelet front-end consistently improves performance.
On average, MLP improves from $0.075$ ROI and $0.813$ Sharpe to $0.165$ and $1.079$, and LSTM improves from $0.191$ ROI and $1.656$ Sharpe to $0.317$ and $1.879$.
WaveLSFormer achieves the best overall ROI and Sharpe with virtually the same parameter count and slightly fewer FLOPs than the vanilla Transformer.
It also surpasses Wavelet+LSTM, improving ROI from $0.317$ to $0.607$ and Sharpe from $1.879$ to $2.157$.
These gains without disproportionate scaling suggest that the improvement is driven by the inductive bias of the neural wavelet front-end and its integration with the Transformer backbone, establishing a new Pareto frontier of performance versus complexity.

\begin{table}[!h]
\centering
\setlength{\tabcolsep}{2pt}
\scriptsize
\caption{Overall average performance and model complexity of all backbones with and without the wavelet front-end. Numbers are averaged over 6 industry groups and 10 random seeds. Model complexity is reported as parameter count and FLOPs per forward pass. Best ROI and Sharpe are in \textbf{bold}, second best are \underline{underlined}.}
\label{tab:overall_baselines}
\renewcommand{\arraystretch}{0.95}
\resizebox{0.92\linewidth}{!}{%
\begin{tabular}{lccccc}
\toprule
Model & Params (M) & FLOPs (G) & Avg. ROI & Avg. Sharpe \\
\midrule
MLP                & 8.146 & 3.468   & 0.075$\pm$0.023 & 0.813$\pm$0.311 \\
MLP + Wavelet      & 8.151 & 10.412  & 0.165$\pm$0.045 & 1.079$\pm$0.139 \\
LSTM               & 12.538 & 492.614 & 0.191$\pm$0.029 & 1.656$\pm$0.495 \\
LSTM + Wavelet     & 13.298 & 530.862 & \underline{0.317$\pm$0.049} & \underline{1.879$\pm$0.221} \\
Transformer        & 15.928 & 665.921 & 0.225$\pm$0.056 & 1.024$\pm$0.122 \\
WaveLSformer (ours) & 15.943 & 659.742 & \textbf{0.607$\pm$0.045} & \textbf{2.157$\pm$0.166} \\
\bottomrule
\end{tabular}%
}
\end{table}

\section{Conclusion}
\label{sec:conclusion}
In this paper, we proposed \emph{WaveLSFormer}, a learnable wavelet-based Transformer for intraday long-short equity trading that combines a neural wavelet front-end with a Transformer backbone and outputs continuous long/short positions under a fixed risk budget.
WaveLSFormer optimizes a trading-aware objective that couples soft-label supervision with ROI and Sharpe oriented regularization, targeting risk-adjusted returns rather than point-wise forecasting accuracy.
Across five years of hourly U.S.\ equity data from 29.10.2020 to 29.10.2025 over six industry groups and ten random seeds, WaveLSFormer consistently outperforms matched MLP, LSTM, and Transformer baselines with and without wavelet front-ends.
Overall, WaveLSFormer achieves $0.607 \pm 0.045$ ROI and $2.157 \pm 0.166$ Sharpe, while the plain Transformer attains $0.225 \pm 0.056$ ROI and $1.024 \pm 0.122$ Sharpe.
Across backbones, the neural wavelet module improves profitability with modest computational overhead, and ablations confirm that learnable filters and gated cross-frequency injection drive the gains rather than backbone capacity alone.

\subsection{Limitations}
Our evaluation uses idealized trading assumptions and does not model transaction costs, slippage or market impact, bid-ask spreads, or leverage and turnover constraints.
We also study a subset of U.S.\ industries at a single bar frequency, so generalization to other universes, liquidity regimes, and market conditions requires further validation.

\subsection{Future work}
We will incorporate differentiable cost and constraint models to enable constrained optimization during training, and explore downside-risk-aware objectives beyond Sharpe-style regularization.
We also plan to study regime-adaptive and online learning, and evaluate the framework on broader asset universes and longer histories.

\newpage

\bibliographystyle{plainnat}
\bibliography{references}

\end{document}

\let\WriteBookmarks\relax
\def\floatpagepagefraction{1}
\def\textpagefraction{.001}

\shorttitle{Supplementary material}
\shortauthors{Li et~al.}

\title[mode=title]{Supplementary Material for A Learnable Wavelet Transformer for Long-Short Equity Trading and Risk-Adjusted Return Optimization}

\author[1]{Shuozhe Li}
\fnmark[1]
\ead{shuozhe.li@utexas.edu}

\author[2]{Du Cheng}
\fnmark[1]
\ead{v6hit7cd@gmail.com}

\author[1,3]{Leqi Liu}
\cormark[1]
\ead{leqiliu@utexas.edu}

\affiliation[1]{organization={Department of Computer Science, The University of Texas at Austin},
            city={Austin},
            state={TX},
            postcode={78712},
            country={USA}}

\affiliation[2]{organization={College of Information Science and Engineering, Northeastern University},
            city={Shenyang},
            postcode={110819},
            country={China}}

\affiliation[3]{organization={Information, Risk and Operations Management Department, The University of Texas at Austin},
            city={Austin},
            state={TX},
            postcode={78712},
            country={USA}}

\cortext[cor1]{Corresponding author}
\fntext[fn1]{Shuozhe Li and Du Cheng contributed equally to this work.}

\begin{abstract}
This supplementary material provides additional industry-level trading performance curves.
\end{abstract}

\maketitle

\section*{Supplementary Figures}

\renewcommand{\thefigure}{S\arabic{figure}}
\setcounter{figure}{0}

\begin{figure}[H]
  \centering
  \includegraphics[width=\linewidth]{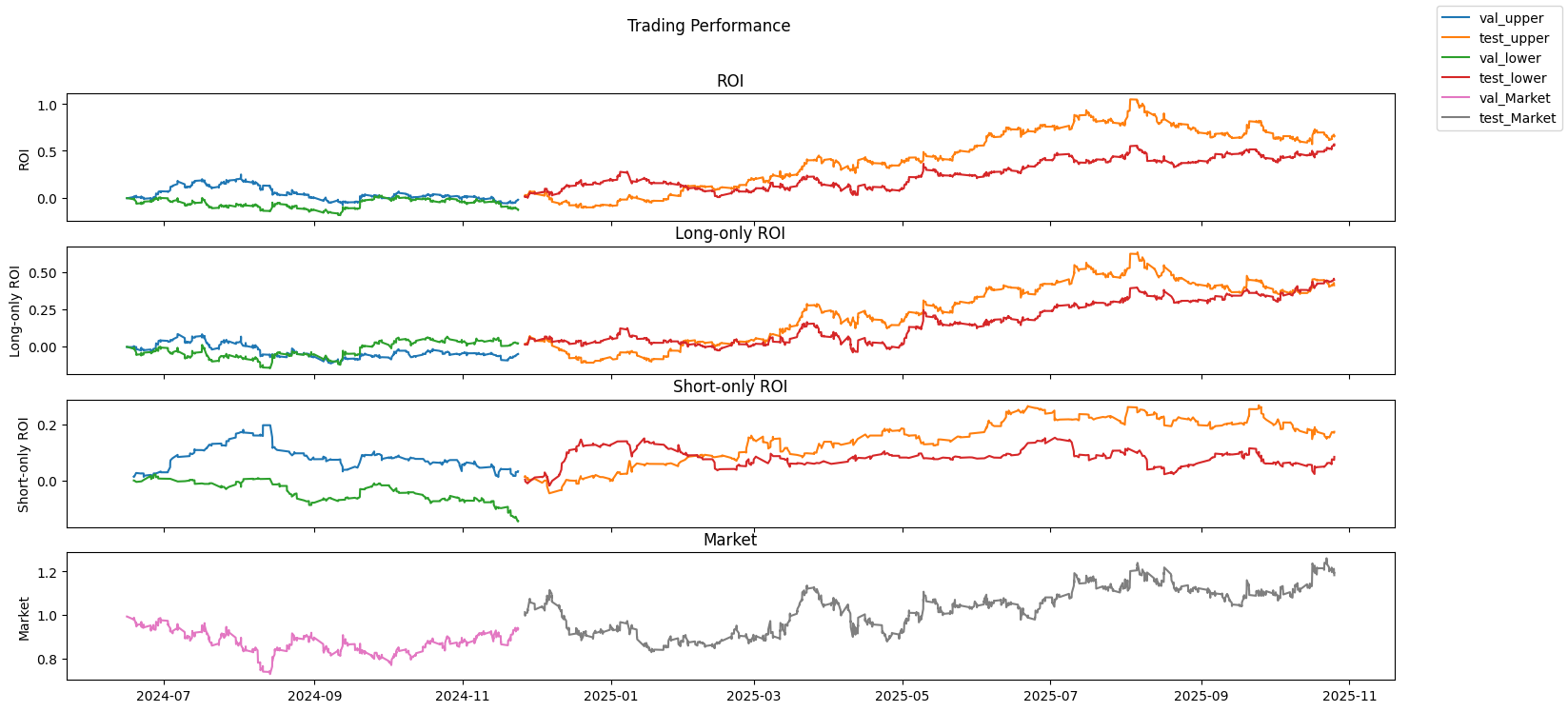}
  \caption{Biotechnology: trading performance curve.}
  \label{fig:s_biotechnology}
\end{figure}

\begin{figure}[H]
  \centering
  \includegraphics[width=\linewidth]{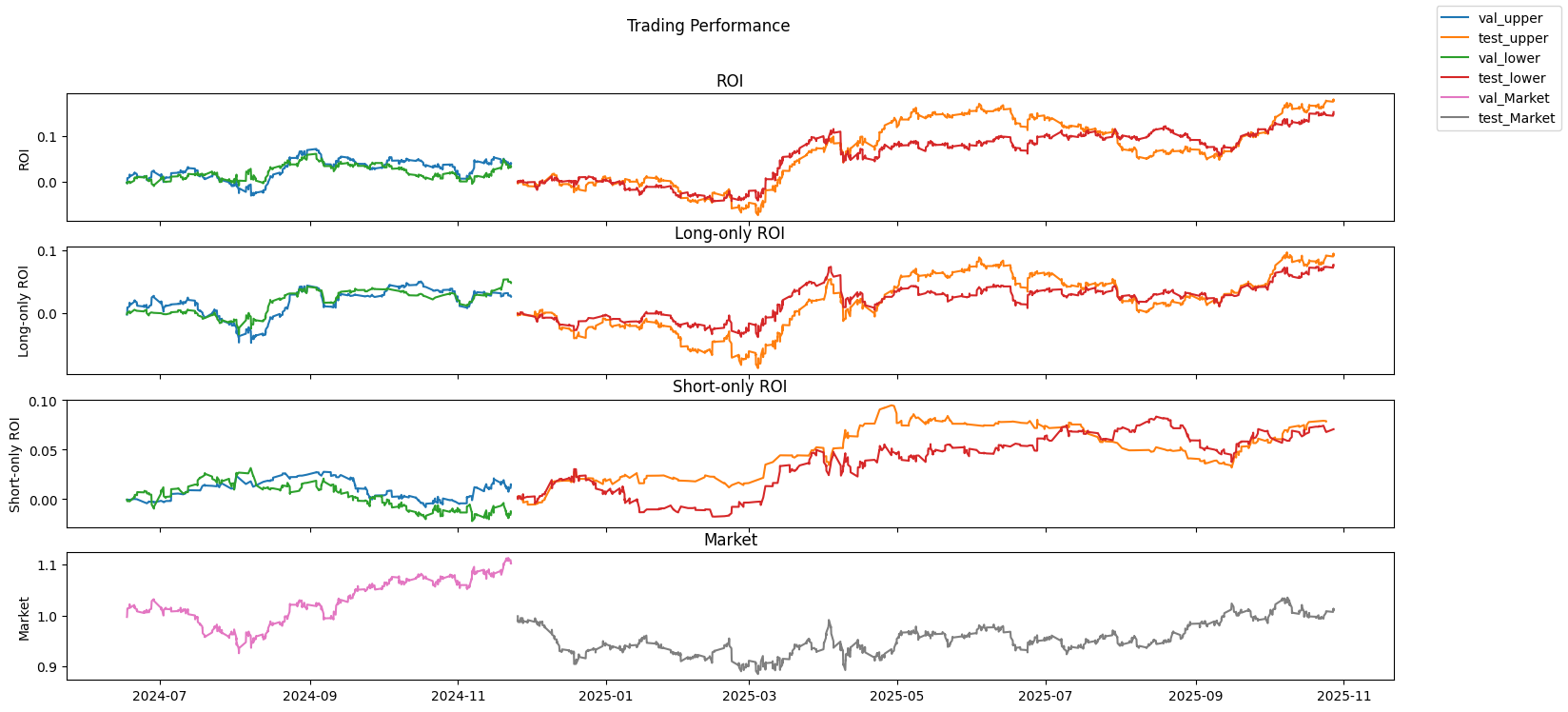}
  \caption{Life Insurance: trading performance curve.}
  \label{fig:s_life_insurance}
\end{figure}

\begin{figure}[H]
  \centering
  \includegraphics[width=\linewidth]{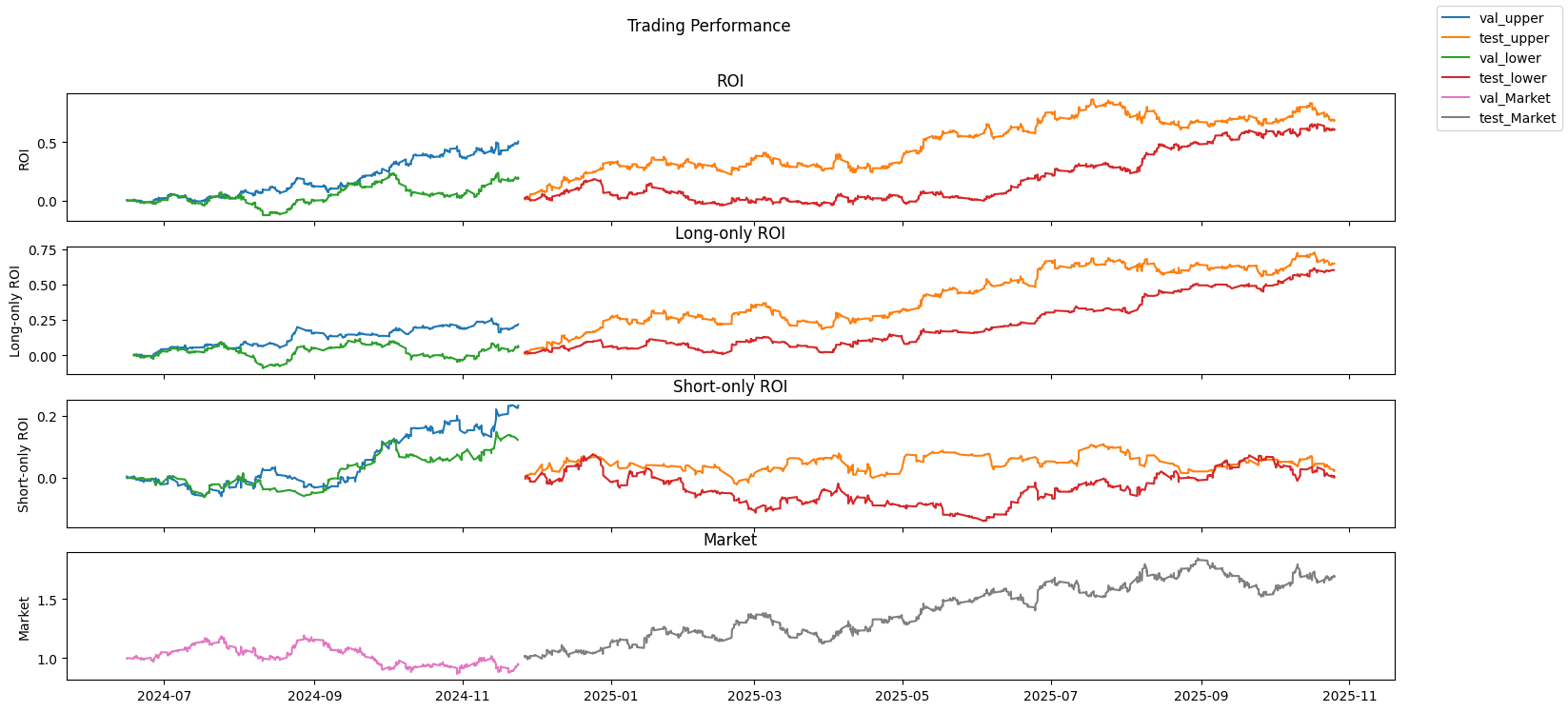}
  \caption{Medical Devices: trading performance curve.}
  \label{fig:s_medical_devices}
\end{figure}

\begin{figure}[H]
  \centering
  \includegraphics[width=\linewidth]{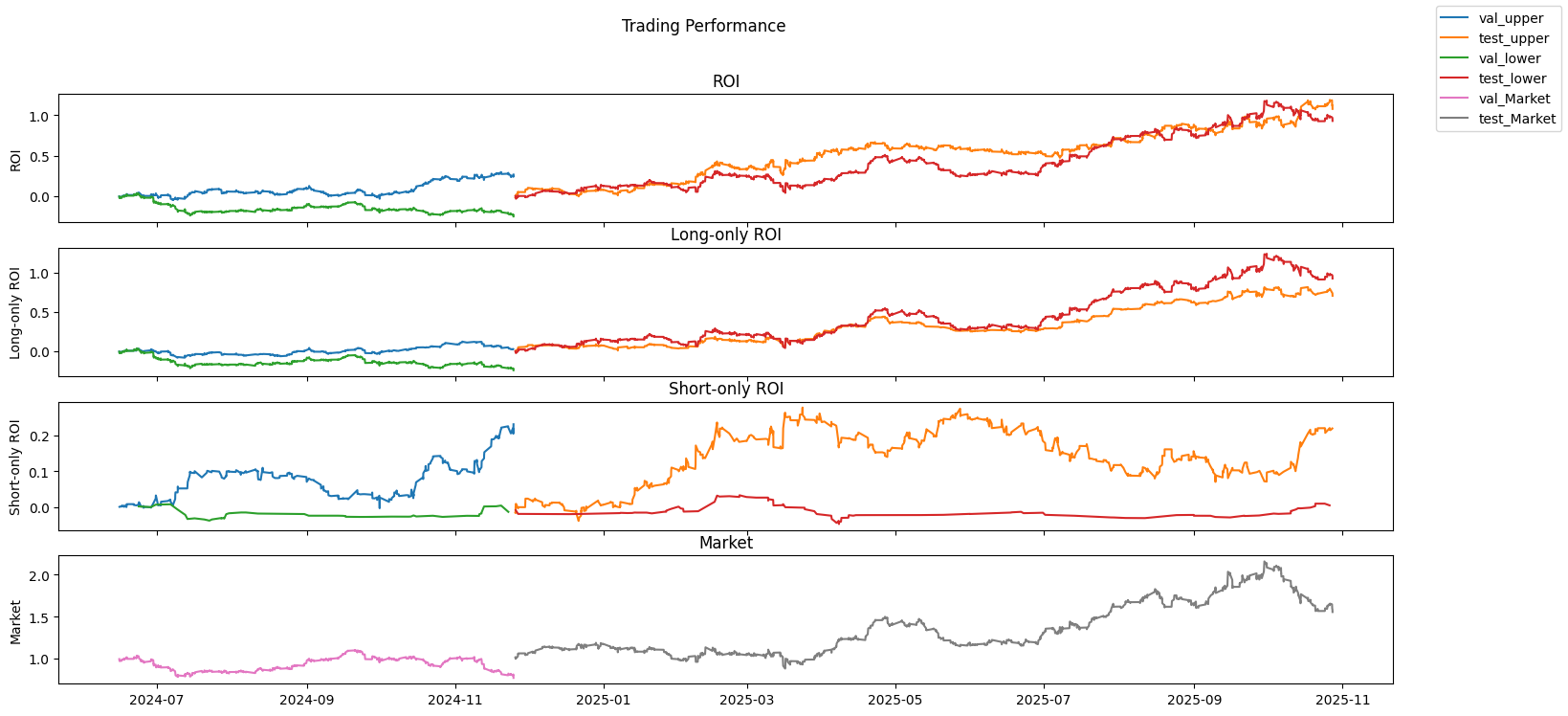}
  \caption{Semiconductor: trading performance curve.}
  \label{fig:s_semiconductor}
\end{figure}